\documentclass{article}
\PassOptionsToPackage{numbers, sort, compress}{natbib}

\usepackage{hyperref}       \usepackage{url}            \usepackage{booktabs}       \usepackage{amsfonts}       \usepackage{amsmath}
\usepackage{amssymb}
\usepackage{amsthm}         \usepackage{nicefrac}       \usepackage{microtype}      \usepackage{graphicx}       \usepackage{mwe}            \usepackage{paralist}       \usepackage{comment}        \usepackage{subcaption}     \usepackage{tcolorbox}
\usepackage{authblk}
\usepackage[british]                         {babel}
\usepackage[capitalise, nameinlink, noabbrev]{cleveref}
\usepackage[inline]                          {enumitem}
\usepackage{booktabs}       \usepackage{multicol}       \usepackage{multirow}       \usepackage{wrapfig}
\usepackage{color}          \usepackage{soul}           \usepackage{siunitx}
\usepackage{thmtools} 
\usepackage{thm-restate}
\usepackage{xspace}
\usepackage{bbold}
\usepackage{xcolor}
\usepackage{bm}
\usepackage[hang, flushmargin]{footmisc} \usepackage{titletoc}

\crefname{equation}{Eq.}{Eqs.}
\usepackage[main, final]{neurips_2025}

\author[1,2,3]{Ernst R\"oell}
\author[1,2]{Bastian Rieck}

\affil[1]{AIDOS Lab, University of Fribourg, Switzerland}
\affil[2]{Institute of AI for Health, Helmholtz Munich, Germany}
\affil[3]{Technical University of Munich, Germany}

\graphicspath{{figures/}}

\newtcolorbox{framed}[1][]{left         = 2.00mm,
  right        = 2.00mm,
  boxsep       = 0.00mm,
  colback      = black!10,
  colframe     = black!50,
  sharp corners,
  #1
}

\definecolor{bleu}     {RGB}{ 49, 140, 231}
\definecolor{airlanebleu} {RGB}{ 25, 69, 219}
\definecolor{lightgrey}{RGB}{230, 230, 230}

\hypersetup{colorlinks  = true,
  urlcolor    = airlanebleu,
  linkcolor   = airlanebleu,
  citecolor   = airlanebleu,
}

\DeclareMathAlphabet{\mathbbold}{U}{bbold}{m}{n}

\newcommand*{\boldone}{\mathbbold{1}}

\newcommand{\ect}{ECT\xspace}
\newcommand{\encoder}{IP-Encoder\xspace}
\newcommand{\downsampler}{IP-Downsampler\xspace}

\newcommand{\vae}{IP-VAE\xspace}

\newcommand{\shapenet}{ShapeNet\xspace}
\newcommand{\pointflow}{PointFlow\xspace}
\newcommand{\ectsmall}{IPT-64\xspace}

\newcommand{\ipt}{IPT\xspace}

\title{Point Cloud Synthesis Using\\ Inner Product Transforms}

\begin{document}

\maketitle

\begin{abstract}
  Point cloud synthesis, i.e.\ the generation of novel point clouds from
  an input distribution, remains a challenging task, for which numerous
  complex machine learning models have been devised.
We develop a novel method that encodes geometrical-topological
  characteristics of point clouds using inner products, leading to a
  highly-efficient point cloud representation with provable expressivity
  properties.
Integrated into deep learning models, our encoding exhibits high
  quality in typical tasks like reconstruction, generation, and
  interpolation, with inference times orders of magnitude faster than
  existing methods.
\end{abstract}
 
\section{Introduction} \label{sec:Introduction}

Point clouds are a data modality of crucial relevance for numerous
domains. While computer graphics is the predominant application area,
where point clouds are often used as precursor to more structured
representations like meshes, they also occur in higher dimensions in
the form of sensor data, for instance.
However, the synthesis of hitherto-unseen point clouds from a given
distribution still proves to be a challenging task, with numerous models
aiming to address it~\citep{Xu23a}.
The complexity arises because of the sparsity and `set-like'
structure of point clouds, making it hard to generalise existing
machine learning architectures directly.
State-of-the-art methods thus typically require large amounts of
compute, exhibiting long training and inference times.
With recent work~\citep{Tancik20a} demonstrating that a change of
perspective---like exchanging raw coordinates for
Fourier-based features---can make comparatively simple deep-learning
architectures competitive in computer-vision tasks, our paper explores
the question to what extent novel representations of point clouds can
lead to gains in computational performance \emph{without} sacrificing
too much quality.

To obtain such representations, we build on a multi-scale
geometrical-topological descriptor~\citep{Turner14a} based on
\emph{inner products} of coordinates representing a high-dimensional
shape.
Using suitable approximations, this descriptor permits us to represent
a point cloud as a \emph{single} 2D image.
Unlike other single-view representations, however, the coordinates of
the image represent a different `domain,' namely geometrical-topological
aspects.
In fact, this mapping is computationally efficient and \emph{injective},
making it theoretically possible to reconstruct a point cloud from its
descriptor~(we will demonstrate that this property also holds empirically
when working with approximations).
We refer to this mapping as the \emph{Inner Product Transform}~(\ipt) and use
it for the generation of 3D point clouds.\footnote{
	For readers familiar with computational topology, the \ipt is
	a special case of the \emph{Euler Characteristic Transform}~(\ect).
	Our focus on point clouds permits us to formulate the \ipt without
	any background knowledge in computational topology, while also
	resulting in substantially simplified proofs and more powerful
	statements about its theoretical properties.
}
Our work primarily focuses on building highly-efficient models that
enable real-time inference in settings of limited computational
resources while also being trainable on commodity hardware.

Our paper is built on \emph{two core notions}, the first one being that we
treat point cloud generation as a two-step task, with the first step
being an image generation task, yielding a two-dimensional
descriptor, followed by a~(multi-modal) image-to-point-cloud
\emph{reconstruction} task.
We realise the two steps using separate machine learning models, thus
substantially decreasing architectural complexity---in effect, we
represent a point cloud as a special image that can be \emph{generated},
providing a bridge to the point cloud domain.
The second core notion is that the generated image is a faithful, i.e.\
\emph{injective}, representation of the point cloud, indicating that
it is possible to perfectly reconstruct a point cloud from
its descriptor.
Our paper makes the following \textbf{contributions}:
\begin{compactenum}
	\item We present a novel generative image-to-point-cloud pipeline 
	based on inner products that allows training and inference times to 
	be \emph{orders of magnitude faster} while retaining high 
	quality generation.
\item We show that our representation yields a stable latent space,
	which permits
\begin{inparaenum}[(i)]
		\item high-quality interpolation tasks, and
		\item solving different out-of-distribution tasks \emph{without} the
		need for retraining while still maintaining high quality.
	\end{inparaenum}
\item We demonstrate injectivity and other advantageous
	properties of our descriptor, allowing the generalisation of our
	method to point clouds of arbitrary dimensions.\end{compactenum}
Our code is available at \url{https://github.com/aidos-lab/inner-product-transforms}
and is released under a BSD-3-Clause license.
 \section{Related Work} \label{sec:Related Work}
Point clouds being a nigh-ubiquitous data modality, numerous models
already exist to tackle classification or generation tasks~\citep{Xu23a}.
The lack of structure, as well as the requirement of permutation
invariance, imposes constraints on the underlying
computational architecture, typically substantially increasing model
complexity~\citep{Qi2017, zaheer2017deep}.
To solve generation tasks, many methods opt for \emph{jointly} learning
the generation of the shape, i.e.\ the surface or object the points are
sampled from, as well the mapping of points onto that object.
This core idea drives several recent state-of-the-art models, including
Point-Voxel CNN~\citep{liu2019point}, PointFlow~\citep{pointflow},
SoftFlow~\citep{kim2020softflow}, Point Voxel
Diffusion~\citep[PVD]{zhou2021pvd} and LION~\citep{Vahdat2022a}.
While these models exhibit high-quality results, their architectures
require long training and inference times.

By contrast, our approach only needs to model a distribution of
geometrical-topological descriptors, represented as 2D images, from
which we subsequently \emph{reconstruct} a point cloud again.
While reconstructing point clouds from images is an active field of
research~(see \citet{Fahim2021a} for a survey on single-view
reconstruction), such images are
typically depth images or snapshots taken from a specific
position around the object.
As such, they are not necessarily yielding a faithful, unique
representation of an object.
Another close analogue to our method is given by `structure-from-motion'
approaches~\citep{Oezyesil17a}, which reconstruct complex geometries
based on \emph{sets} of images, taken from different spatial viewpoints.
For our method, however, the viewpoints are represented as unit vectors
on a sphere, which are used to `probe' the point cloud from a specific
direction.

Hence, a crucial property of our method, the \emph{Inner Product
Transform}~(\ipt) is that it theoretically yields a faithful
representation, hence permitting an injective mapping between the image
domain and the shapes we aim to reconstruct~(or generate).
We observe this property to hold empirically when working with
discrete approximations.
This is due to the fact that the \ipt is a special case of a general
geometrical-topological descriptor, the \emph{Euler Characteristic
    Transform}~(\ect), which studies shapes at multiple scales and from
multiple directions, providing a unique characterisation~\citep{Turner14a}.
Being a stable~\citep{Dlotko24a, George25a} and efficient descriptor, the \ect
is often used to solve questions in data science, mostly in the form of
`hand-crafted' features for classification and regression
tasks~\citep{amezquita2022measuring, Crawford20a, Marsh24a,
    nadimpalli2023euler, Munch23a, Toscano25a, vonRohrscheidt25a}.
Recent work addressed this shortcoming and enabled the use of the ECT in
machine-learning applications in the form of a generic
differentiable \emph{computational layer}~\citep{Rieck25a, Roell24a} or a \emph{positional
encoding}~\citep{Amboage25a}.
Notably, the \ect remains a unique characterisation even
when using a finite number of directions~\citep{curry2022many,
    Ghrist18a}, meaning that, theoretically, it can be inverted to reconstruct
the input data.
Practically, however, inversion is presently possible only for select data
modalities like planar graphs~\citep{Fasy18a}.
\textbf{While restricted to point clouds, our method is thus the first to enable the
inversion of such a descriptor for input data of arbitrary
dimensionality, making it possible to use it in the context of
generative models.}
 \section{Methods} \label{sec:Methods}
Our method, the \emph{Inner Product Transform}~(\ipt), can be
intuitively understood as a filtering process of point clouds, employing
different sets of hyperplanes, created by a set of
\emph{directions}~(i.e.\ normal vectors). The inner products of
point cloud coordinates with a direction vector are then used to
parametrise a \emph{curve} that counts the number of points below the
hyperplane. By stacking curves, we obtain a 2D image representation of
the point cloud.
The \ipt is a special case of the \emph{Euler Characteristic
	Transform}~\citep[ECT]{Turner14a}, but our subsequent description is
self-contained, does not require any knowledge of topology, and presents
simplified proofs of all properties.

\begin{figure*}[tbp]
    \center
    \includegraphics[width=\textwidth]{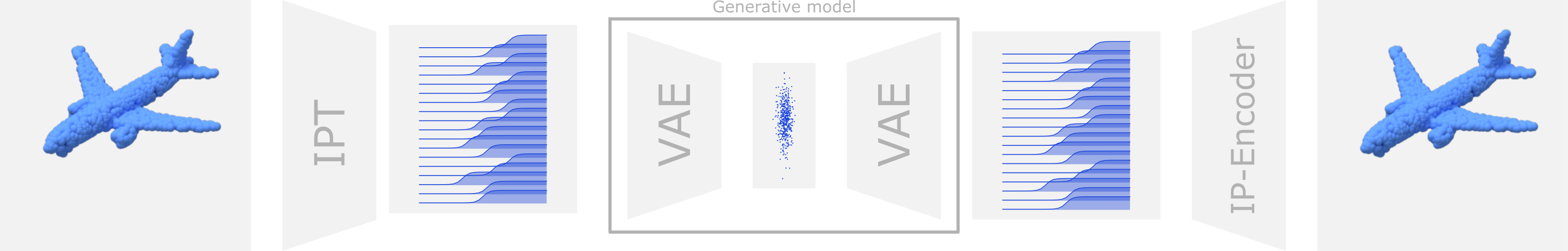}
    \caption{Given a point cloud on the left, we compute its \emph{Inner Product
        Transform}~(\ipt).
For generative tasks, we train a generative model~(middle) to
        reconstruct and generate the distribution of \ipt{}s.
The (possibly-generated) \ipt is then passed through the encoder model to
        obtain the reconstructed (or novel) point cloud.
Our pipeline is decoupled, permitting \emph{any} generative image model
        to be used to generate point clouds.
    }
    \label{fig:overview}
\end{figure*}
 
\subsection{Inner Product Transforms}
Let $S^{n-1}$ denote the unit sphere in $\mathbb{R}^n$.
Given a point cloud $X\subset \mathbb{R}^{n}$, a fixed direction vector $\xi\in
	S^{n-1}$, and a \emph{height} $h\in \mathbb{R}$, we define the set
$X_{\xi,h} := \{ x \in X \mid \langle x, \xi \rangle \leq h \}$, where
$\langle x, \xi \rangle$ is the Euclidean dot product.
The set $X_{\xi,h}$ contains all points below the hyperplane spanned by
$\langle x, \xi \rangle =h$, and we denote its cardinality by
$\chi(X_{\xi,h})$.\footnote{For readers familiar with topology, this notation is an allusion to
	the \emph{Euler Characteristic}.
}
We then define the \emph{Inner Product Transform} as
\begin{equation}
	\begin{aligned}
		\mathrm{IPT}(X)\colon S^{n-1}\times \mathbb{R} & \to \mathbb{N}            \\
		(\xi, h)                                       & \mapsto \chi(X_{\xi,h}).
	\end{aligned}
	\label{eq:IPTEquation}
\end{equation}
A point $x\in X$ is included in $X_{h,\xi}$, thus affecting
$\chi(X_{\xi,h})$, if and only if its height $h_x := \langle x,\xi
	\rangle$ along $\xi$ is less than $h$.
We can thus formulate the contribution of a point~$x$ to the \ipt along
each direction in terms of an \emph{indicator function}:
\begin{equation}
	\boldone_{x}(\xi, h) :=
	\begin{cases}
		1 & \text{if } \langle \xi, x \rangle \leq h \\
		0 & \text{otherwise}.
	\end{cases}
\end{equation}
This enables us to rewrite \cref{eq:IPTEquation} as
\begin{equation}
	\begin{aligned}
		\mathrm{IPT}(X)\colon S^{n-1}\times \mathbb{R} & \to \mathbb{N}                              \\
		(\xi, h)                                       & \mapsto \sum_{x\in X} \boldone_{x}(\xi,h).
	\end{aligned}
	\label{eq:Indicator}
\end{equation}
Following ideas from \citet{Roell24a}, we can replace all indicator
functions with sigmoid functions, i.e.\
\begin{equation}
	\begin{aligned}
		\widehat{\mathrm{IPT}}(X)\colon S^{n-1}\times \mathbb{R} & \to \mathbb{R}                                                 \\
		(\xi, h)                                                 & \mapsto \sum_{x\in X} S(\lambda(\langle \xi, x \rangle - h)),
	\end{aligned}
	\label{eq:IPTIndicator}
\end{equation}
where $\lambda$ denotes a scale parameter, which controls how closely
the sigmoid function approximates the indicator
function~(see \cref{app:Additional Experiments} for
additional ablations demonstrating the stability of this approach).
In practice, we sample~$n_d$ directions and discretise all heights
with~$n_h$ steps, thus representing $\widehat{\mathrm{IPT}}$ as an
\emph{image} of resolution $n_h \times n_d$.

\paragraph{Properties.}
The \ipt and its approximation $\widehat{\mathrm{\ipt}}$ from
\cref{eq:IPTIndicator} satisfy several important properties. We first
focus on the case of the \ipt and note that it is \emph{injective},
i.e.\ a point cloud can be perfectly reconstructed from its descriptor.
\textbf{Unlike existing injectivity results~\citep{Turner14a, Ghrist18a}, which
	focus on geometric simplicial complexes, we prove that for point clouds
	in $\mathbb{R}^n$, it is sufficient to use $n+1$ affinely-independent
	directions for the reconstruction.}
\begin{restatable}{theorem}{iptinjective}
	\label{thm:iptinjective}
	Given two point clouds $X, Y$ with $X \neq Y$, we have 
	$\mathrm{\ipt}(X) \neq \mathrm{\ipt}(Y)$.
\end{restatable}
An important consequence of injectivity is that it allows us to
formulate a metric for the set of point clouds contained in a ball of
fixed radius.
\begin{restatable}{lemma}{iptmetric}
	\label{lem:iptmetric}
	Let $B_n(R)$ denote a ball of radius $R$ in $\mathbb{R}^n$. For two
	point clouds $X, Y \subset B_n(R)$, define their \emph{distance} as
\begin{equation}
		\begin{aligned}
			d(X,Y) := \frac{1}{\vert \Xi \vert }\sum_{\xi\in\Xi}\Vert
			\mathrm{IPT}(X)_{\restriction{\xi}}-
			\mathrm{IPT}(Y)_{\restriction{\xi}} \Vert_{2},
		\end{aligned}
	\end{equation}
where $\Xi$ is a finite set of directions and $\Vert\cdot\Vert_{2}$
	is the $L^{2}$-norm restricted to the interval $[-R,R]$. The
	function $d(\cdot,\cdot)$ satisfies the definition of a
	metric.
\end{restatable}
When approximating the \ipt via \cref{eq:IPTIndicator} and a finite
number of directions in practice, the metric defined in
\cref{lem:iptmetric} corresponds to the pixel-wise mean squared error
between discretised \ipt{}s.
Hence, given a sufficient number directions, we can formulate
a \emph{loss function} based on this metric, which admits
a highly-efficient implementation and will turn out to lead to
high-quality results.
We also obtain a result that permits us to calculate \ipt{}s from
disjoint unions of point clouds.
\begin{restatable}{lemma}{iptlinear}
	Let $X,Y\subset\mathbb{R}^{n}$ be disjoint point clouds, then
	\begin{equation}
		\mathrm{IPT}(X\cup Y) = \mathrm{IPT}(X) + \mathrm{IPT}(Y).
	\end{equation}
\end{restatable}
Finally, we can prove that the \ipt is \emph{surjective} on convex
linear combinations.
\begin{restatable}{theorem}{iptsurjective}
	\label{thm:iptsurjective}
	Given two point clouds $X,Y\subset \mathbb{R}^{n}$,
	the \ipt is \emph{surjective} for the rational linear
	subspace spanned by $\mathrm{IPT}(X)$ and $\mathrm{IPT}(Y)$, up to a
	rational scaling factor.
In particular, for $p,q\in \mathbb{N}_{0}$ with $0\leq p \leq q$ and 
	$q > 0$ we have
	\begin{equation}
		\frac{p}{q}\mathrm{IPT}(X) + \frac{q-p}{q}\mathrm{IPT}(Y) = \frac{1}{q}\mathrm{IPT}\big( Z \big),
	\end{equation}
where $Z = \cup_{p} X  \cup_{q-p} Y$.
\end{restatable}
A direct consequence of \cref{thm:iptsurjective} is that, along the
\emph{linear interpolation} between two \ipt{}s, there are only valid
\ipt{}s, i.e.\ each interpolation step affords a perfect reconstruction
in theory. We analyse this aspect further in \cref{sec:Interpolation}.
\textbf{We note that most of these properties~(except injectivity) only
	hold for the \ipt but, as we will later demonstrate, $\widehat{\mathrm{\ipt}}$,
	the approximation to the \ipt, retains these properties in practice.}

\subsection{The \encoder}

The existence of \cref{thm:iptinjective} unfortunately does not lead to
a practical algorithm for `inverting' an \ipt. We therefore suggest
an approach based on neural networks and describe the \encoder, which
encodes an \ipt to a point cloud.
Subsequently, this will enable us to learn \emph{distributions} of
\ipt{}s and reconstruct new point clouds in a generative setting and, as
we shall see, even in the setting of out-of-distribution data.
Since our representation is permutation-invariant, our \encoder model
directly inherits this invariance, which substantially reduces the
complexity of its architecture.

\paragraph{Model architecture.}
Given the structure of the \ipt as an image, CNN architectures provide
suitable base models. However, since the direction vectors $\xi \in
	S^{n-1}$ cannot be consistently ordered along one dimension, we need to
reframe the input data. Specifically, we consider an \ipt, normalised to
$[-1,1]$, as a \emph{multi-channel, one-dimensional signal}.
Our \encoder model then consists of multiple 1D convolutional layers
followed by fully-connected layers, resulting in a conceptually simple
and efficient architecture.
To \emph{generate} a novel \ipt, we can use any generative model for
images. In our experiments, we will use an architecture based on
a convolutional variational autoencoder, denoted by \vae~(see
\cref{app:Architectural Details} for more architectural
details).

\paragraph{The \ipt as an optimisation problem.}
As an alternative to the machine learning model above, we also
investigate whether the inversion of an \ipt can be turned into
an \emph{optimisation problem}.
Due to the differentiability of our approximation
$\widehat{\mathrm{\ipt}}$ and given a known \ipt,
minimising a loss function, such as the one described in
\cref{lem:iptmetric}, using backpropagation should result in a suitable
approximation of the unknown point cloud one wishes to recover.
However, while this works in \emph{theory}, it requires the input and
target to be a `correct' $\widehat{\mathrm{\ipt}}$ of a point cloud. In
practice, this is \emph{not} guaranteed: Generative models may sample from
their latent spaces and output samples that are \emph{close} to being an
$\widehat{\mathrm{\ipt}}$ of a point cloud without actually satisfying all structural
constraints. To some extent, we argue that this is even the desired
behaviour since a generative model would otherwise just rehash its
inputs~\citep{Kamb25a}.
Hence, any optimisation-based method will not necessarily result in
realistic reconstructions in a generative setting. We nevertheless
investigate such an optimisation procedure in an ablation
study in the appendix~(\cref{fig:shapenetcore_rendered}).

\paragraph*{Topological loss functions.}
The fact that \cref{lem:iptmetric} shows the \ipt to be a metric
naturally poses the question if its approximation
$\widehat{\mathrm{IPT}}$ can be used as an effective and efficient loss
term.
While in principle the $\widehat{\mathrm{\ipt}}$ is sufficient as a loss
term for training a point cloud reconstruction model, training time can
be reduced through a combination of a low-resolution ($64\times 64$)
$\widehat{\mathrm{\ipt}}$ combined with the Chamfer Distance~(CD).
The CD is a fast-to-compute (pseudo-)metric often used in point cloud
evaluation~(see \cref{app:Metrics for Point Clouds} for more details).
The $\widehat{\mathrm{\ipt}}$ loss ensures that the overall geometry and point cloud density
of the object is captured, whereas the CD loss ensures that fine-grained
details are accounted for.
Building a joint loss, combining the $\widehat{\mathrm{\ipt}}$ and CD,
thus results in a density-aware loss term that takes the global density
of a point cloud into account.

\paragraph*{Latent space.}
Interpolation between samples provides valuable insight into the
capacities of our generative model.
\cref{thm:iptsurjective} provides an intuition for interpolation using
$\widehat{\mathrm{\ipt}}$s.
To this end, suppose we perform linear interpolation between point
clouds~$X, Y$, using a parameter~$p$ over the interval $[0,1]$, which is
partitioned in $q$ equidistant steps.
Then, at step $p / q$, the point cloud $Z=\cup_{p} X  \cup_{q-p} Y$ is
the union of $p$ copies of $X$ and $q-p$ copies of $Y$.
During the interpolation, the number of copies of $X$ is increased and
the number of copies of $Y$ is reduced.
We may interpret this intuitively as transporting `mass'
from $Y$ to $X$ over the course of $q$ steps.
Our \encoder model averages all copies during reconstruction in a natural
fashion, resulting in smooth transitions between point clouds.
 \section{Experiments} \label{sec:Experiments}

We demonstrate the effectiveness, efficiency, and overall utility of our
\encoder through a comprehensive suite of experiments.
{
		\bfseries
		Subsequently, to simplify the notation, we will only refer to the
		\ipt, with the understanding that we are calculating an
		\emph{approximation} of it according to \cref{eq:IPTIndicator}.
	}
Throughout our experiments, the emphasis and motivating questions regard 
the expressivity of the \ipt and its capacity to faithfully and 
\emph{effectively} represent point clouds in a practical setting. 
Good results with a minimal architecture underpin the fact the \ipt is an 
effective representation as it shows that the data distribution is easy 
to learn.
Among other things, we show that our \encoder model can
\begin{inparaenum}[(i)]
	\item effectively reconstruct shapes,
	\item create novel point clouds from \emph{generated} \ipt{}s, and
	\item effectively downsample point clouds.
\end{inparaenum}

All our experiments rely on a subset of the \shapenet dataset, and we
adopt the preprocessing and evaluation workflow introduced by
\citet{pointflow}.
Each point cloud in the dataset consists of $2048$ points sampled on the
surface of three shape classes~(airplanes, chairs and cars).
We report the \emph{minimum matching distance}~\citep[MMD]{achlioptas2018learning}
based on the Chamfer Distance~(MMD-CD) or the \emph{Earth Mover's
	Distance}~(MMD-EMD) between reconstructed point
clouds~(see \cref{app:Metrics for Point Clouds} for a brief description
of these evaluation metrics).
Unless otherwise mentioned, the reported Chamfer Distance is scaled by
\num{1e4} and the Earth Mover's Distance is scaled by \num{1e3}.
Our hardware consists of an NVIDIA GeForce RTX 4070 with 12GB VRAM and 
an 13th Gen Intel(R) Core(TM) i7-13700K with 32GB RAM.
We compare our methods to several state-of-the-art models, namely
\begin{inparaenum}[(i)]
	\item PointFlow~\citep{pointflow},
	\item SoftFlow~\citep{kim2020softflow},
	\item ShapeGF~\citep{cai2020learning}, 
	\item Canonical VAE~\citep{Cheng2022a} and
	\item LION~\citep{Vahdat2022a}.
\end{inparaenum}

\subsection{Reconstructing Point Clouds}
\label{sec:Reconstructing Point Clouds}

Our first set of experiments assesses 
\begin{inparaenum}[(i)]
	\item the reconstruction quality of the \encoder,
	\item the efficacy of the \ipt as a loss, and
	\item its computational efficiency.
\end{inparaenum}

\paragraph{Architecture and experimental setup.}
We consider the \ipt as a 1D signal, with each direction corresponding to a
channel, sample $128$ directions uniformly from the unit sphere, and discretise
each direction into $128$ steps, thus obtaining an \ipt with a resolution of
$128\times 128$.
Sampling the directions \emph{randomly} is motivated by the lack of
canonical ordering of unit vectors in three and higher
dimensions~(notice that in two dimensions, the angle affords
a parametrisation that results in a natural ordering).
While `pseudo-ordered' directions, for instance via a spiral along the
unit sphere, \emph{might} potentially lead to better results, they come
at the cost of generality, and we thus refrained from doing so in our
experiments.
To improve this for future work, one could use a \emph{positional
	encoding} of the directions; at present, such an ordering is only
implicitly present through the fixed ordering of columns in the image
we use to represent the \ipt.
Given such a 2D representation of an \ipt, our \encoder consists of four 1D
convolutional layers with batch normalisation, max-pooling, and SiLU activation
functions~(cf.\ \cref{app:Architectural Details}).
After mapping the \ipt into a latent space, we apply a final
$3$-layer MLP to predict the final point cloud.
We use ReLU activation functions for the first two layers and a $\tanh$
activation function for the last layer, since $\tanh$ is better suitable
for bounded outputs.
Subsequently, we train the \encoder \emph{separately} for each of the
classes for $5k$ epochs, using a \mbox{CD + \ipt-64} loss, denoting
a weighted sum of the CD and the \ipt with a resolution of $64\times
	64$.

\begin{table*}[t]
	\centering
	\sisetup{
		detect-all              = true,
		table-format            = 2.2(2),
		detect-mode             = true,
		separate-uncertainty    = true,
		retain-zero-uncertainty = true,
		table-align-text-after  = false,
		mode                    = text,
	}\caption{Reconstruction results on three \shapenet classes.
The best-performing model is highlighted in \textbf{bold}, while the
		second-best is shown in \emph{italics}.
Our method consistently ranks second in terms of the EMD,
		which is recognised as the best metric for reconstruction
		quality~\citep{zhou2021pvd}.
	}\label{tab:shapenet_reconstruction}
\let\b\bfseries
	\let\i\itshape
	\footnotesize \begin{tabular}{l
S[table-format=1.2(2)]@{\hspace{6pt}}S[table-format=1.2(2)]S[table-format=2.2(1.2)]@{\hspace{6pt}}S[table-format=2.2(1.2)]S[table-format=1.2(1.2)]@{\hspace{6pt}}S[table-format=2.2(1.2)]}
		\toprule
		                & \multicolumn{2}{c}{\emph{Airplane}}
		                & \multicolumn{2}{c}{\emph{Chair}}
		                & \multicolumn{2}{c}{\emph{Car}}                                                                                                         \\\midrule
		\textsc{Model}
		                & {\small\sc CD\,($\downarrow$)}
		                & {\small\sc EMD\,($\downarrow$)}
		                & {\small\sc CD\,($\downarrow$)}
		                & {\small\sc EMD\,($\downarrow$)}
		                & {\small\sc CD\,($\downarrow$)}
		                & {\small\sc EMD\,($\downarrow$)}                                                                                                        \\\midrule
		PointFlow       & 1.30 \pm 0.00                       & 5.36 \pm 0.06    & 10.43 \pm 0.02    & 17.54 \pm 0.16    & 6.94 \pm 0.01    & 12.93 \pm 0.19     \\
		SoftFlow        & 1.19 \pm 0.00                       & 4.28 \pm 0.06    & 11.05 \pm 0.03    & 17.68 \pm 0.08    & 6.82 \pm 0.01    & 11.44 \pm 0.10     \\
		ShapeGF         & \i 1.05 \pm 0.00                    & 4.42 \pm 0.04    & \i 5.96 \pm 0.01  & 12.23 \pm 0.11    & 5.68 \pm 0.01    & 9.26 \pm 0.18      \\
		Canonical VAE   & 0.98 \pm 0.0                        & 3.19 \pm 0.03    & 6.56 \pm 0.02     & 8.6 \pm 0.07      & \i 5.44 \pm 0.01 & 6.13 \pm 0.02      \\
LION            & \b 0.3\pm 0.00                      & \b 0.12 \pm 0.00 & \b 0.7   \pm 0.00 & \b 0.14  \pm 0.00 & \b 0.6 \pm 0.00  & \b 0.09   \pm 0.00 \\\midrule
		\encoder (ours) & \i 1.05 \pm 0.0                     & \i 1.57 \pm 0.0  & 9.24 \pm 0.0      & \i 6.19 \pm 0.0   & 5.82 \pm 0.0     & \i 3.18 \pm 0.0    \\\bottomrule
\end{tabular}\end{table*}
 
\begin{figure}[tbp]
	\centering
\begin{subfigure}{.475\textwidth}
		\includegraphics[width=.95\linewidth]{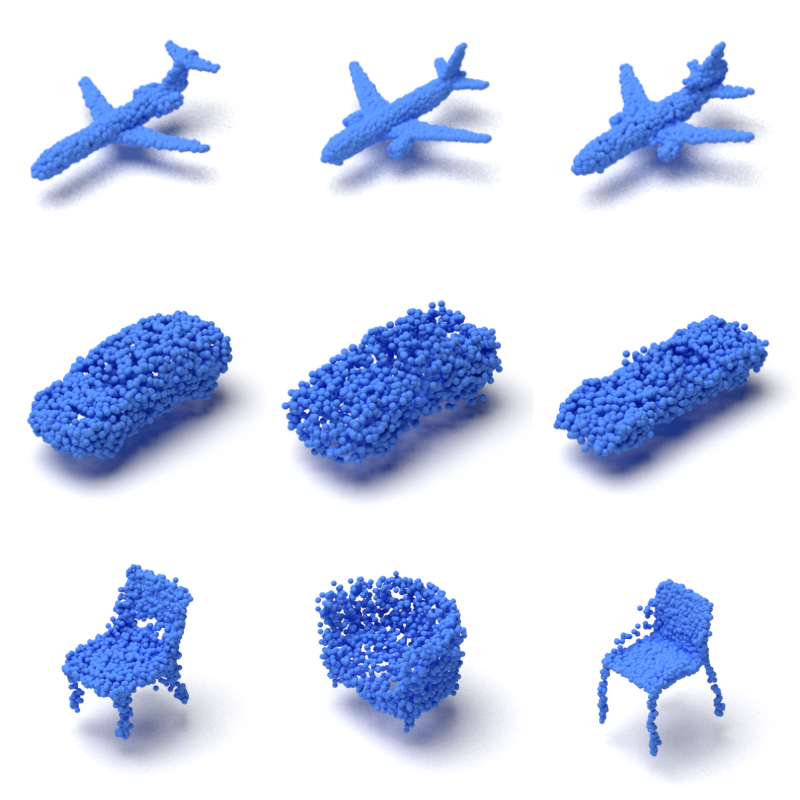}
		\subcaption{\phantom{x}}
		\label{sfig:encoder}
	\end{subfigure}\hfill
\begin{subfigure}{.475\textwidth}
		\includegraphics[width=.95\linewidth]{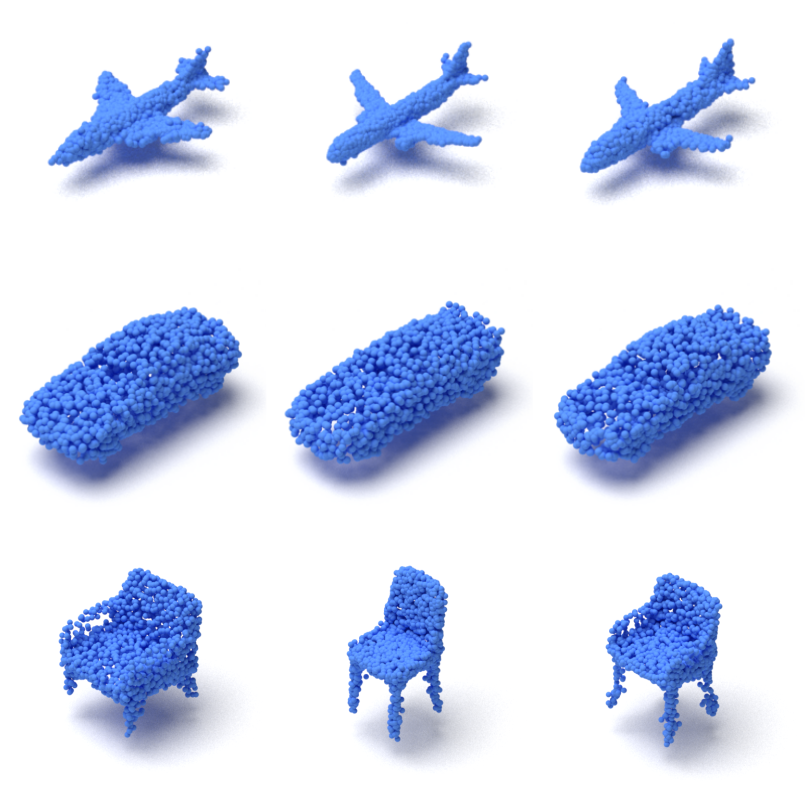}
		\subcaption{\phantom{x}}
		\label{sfig:generated}
	\end{subfigure}
	\caption{Examples of \emph{reconstructed}~\subref{sfig:encoder} and \emph{generated}~\subref{sfig:generated} point clouds using our \encoder model for
		three classes in the \shapenet dataset.
}
	\label{fig:shapenet_both}
\end{figure}

\begin{wraptable}[10]{l}{0.4\linewidth}
	\vspace{-1\baselineskip}
	\centering
	\sisetup{
		detect-all              = true,
		table-format            = 2.2(1.2),
		detect-mode             = true,
		separate-uncertainty    = true,
		retain-zero-uncertainty = true,
	}\caption{Ablation study wrt.\ the loss function. Our joint loss,
		combining CD and \ipt, yields \emph{balanced} results without
		overfitting any of the two metrics.
	}\label{tab:shapenet_reconstruction_ablation}\let\b\bfseries
	\let\i\itshape
	\small \resizebox{\linewidth}{!}{\begin{tabular}{l
S[table-format=1.2]@{\hspace{6pt}}S[table-format=1.2]S[table-format=2.2]@{\hspace{6pt}}S[table-format=2.2]S[table-format=1.2]@{\hspace{6pt}}S[table-format=2.2]}
\toprule
			               & \multicolumn{2}{c}{\emph{Airplane}{\small\,($\downarrow$)}}
			               & \multicolumn{2}{c}{\emph{Chair}{\small\,($\downarrow$)}}
			               & \multicolumn{2}{c}{\emph{Car}{\small\,($\downarrow$)}}                                                                                    \\
			\midrule
			\textsc{Loss function}
			               & {\small CD }
			               & {\small EMD }
			               & {\small CD }
			               & {\small EMD}
			               & {\small CD }
			               & {\small EMD}                                                                                                                              \\
			\midrule
			CD             & 1.00 \pm 0.0                                                & 8.89 \pm 0.0 & 10.44 \pm 0.0 & 32.43 \pm 0.0 & 5.97 \pm 0.0 & 14.96 \pm 0.0 \\
			\ectsmall      & 2.41 \pm 0.0                                                & 1.09 \pm 0.0 & 13.06 \pm 0.0 & 4.29 \pm 0.0  & 7.75 \pm 0.0 & 2.47 \pm 0.0  \\
			\midrule
			CD + \ectsmall & 1.03 \pm 0.0                                                & 1.46 \pm 0.0 & 9.52 \pm 0.0  & 8.44 \pm 0.0  & 6.12 \pm 0.0 & 4.16 \pm 0.0  \\
			\bottomrule
		\end{tabular}}\end{wraptable}
 \paragraph{Results.}
Our reconstructions~(\cref{tab:shapenet_reconstruction} and
\cref{sfig:encoder}) are of high quality, 
despite a comparatively simple architecture.
Our \encoder consistently ranks second in terms of the EMD, known to be the most 
suitable metric to evaluate reconstruction quality~\citep{zhou2021pvd}. 
Notably, our loss term does \emph{not} use the EMD.
Our method also exhibits substantially reduced training times compared
to other methods: The \encoder model requires approximately
\SI{30}{\minute} on a single GPU, compared to \SI{192}{\hour} for
\pointflow and more than of \SI{550}{\hour} for LION.
We additionally perform an ablation study concerning the loss term,
showing that the \emph{combination} of CD and \ipt is crucial for high
reconstruction quality.
As \cref{tab:shapenet_reconstruction_ablation} shows, a joint loss
yields the best quality. In line with prior work, CD on its own tends to
adversely affect point-cloud density~\citep{achlioptas2018learning},
resulting in larger~(i.e.\ \emph{worse}) EMD scores.
Conversely, training only with a \ect loss shows that the reconstructed
follow the density of the underlying object, but may become blurry.
This is partially due to the low resolution of $64$, making the
combination of CD~(for details) and the \ect~(for global density and
structure) crucial.
While it would also possible to use the EMD as a loss term, we find that
our \ectsmall loss function is \emph{substantially faster} than the
EMD~(\SI[round-precision=4]{0.00064}{\second} versus
\SI[round-precision=4]{0.02913}{\second}), making it the preferred loss
function.

\begin{framed}
	We observe the \ipt to be an effective representation of point clouds,
	leading to a conceptually simple model with exceptionally fast
	training times.
\end{framed}

\subsection{Generating Novel Point Clouds}
We now view the \ipt as a latent space, from which we can \emph{sample}
and subsequently \emph{reconstruct} point clouds, demonstrating that
\begin{inparaenum}[(i)]
	\item the \ipt constitutes a stable latent space,
	\item a \emph{distribution} of shapes can be learned through the
	\ipt, and
	\item the \encoder works well in out-of-sample settings.
\end{inparaenum}

\paragraph{Architecture and experimental setup.}
We use \vae, a CNN-style VAE~\citep{Higgins2016betaVAELB} as our generative
model, thus turning point cloud generation into an image generation
task.
Our encoder uses four convolutional blocks with Leaky ReLU
activation functions, batch normalisation, and a final linear layer for
the $64$-dimensional latent space embedding.
We also add additional models, namely
\begin{inparaenum}[(i)]
	\item SetVAE~\citep{kim2021setvae},
	\item Point Voxel Diffusion~\citep[PVD]{zhou2021pvd}, 
	\item Point Straight Flows~\citep{wu2023fast}, and 
	\item XCube~\citep{ren2024xcube}.
\end{inparaenum}
These models lack reconstruction capabilities or the respective
code, so we excluded them from the experiment in
\cref{sec:Reconstructing Point Clouds}.
For training the \vae, we follow the $\beta$-VAE setup, using
KL-divergence and MSE loss terms with $\beta = \num{1e-4}$.
We sample latent vectors from the \vae and consider them to be \ipt{}s,
which we subsequently map to a point cloud using the \encoder model.
Our evaluation of generative performance follows the setup of \citet{pointflow}.

\begin{wraptable}[15]{l}{0.25\linewidth}
	\vspace{-1.0\baselineskip}
	\centering
	\caption{Inference time T in \si{\second} for all methods, measured
		on the same hardware.
Our model is \mbox{orders} of magnitude faster than all others.
	}
	\sisetup{
		detect-all              = true,
		detect-mode             = true,
		table-format            = 2.3,
		round-precision         = 3,
	}\label{tab:performance}
	\let\b\bfseries
	\resizebox{\linewidth}{!}{\begin{tabular}{lcS}
			\toprule
			\textsc{Model} & {\small \textsc{Device}} & {\small\sc {T\,($\downarrow$)}} \\
			\midrule
			PointFlow      & \multirow{5}{*}{GPU}     & .27                             \\
			SoftFlow       &                          & .12                             \\
			ShapeGF        &                          & .34                             \\
			SetVAE         &                          & .03                             \\
			PSF            &                          & 0.04                            \\ 
			PVD            &                          & 29.9                            \\
			\midrule
			\vae           & GPU                      & \b 0.00079                      \\
			\vae           & CPU                      & \b .00588                       \\
			\bottomrule
		\end{tabular}}\end{wraptable}
 \paragraph{Results.}
\cref{tab:shapenet_gen_full} reports numerical results, whereas
\cref{sfig:generated} depicts generated samples from each
class.
We include different quality metrics and note that, despite its
conceptual simplicity, our model consistently ranks among the best two
models in terms of MMD-EMD, and exhibits performance on a par with more
elaborate architectures.
This is noteworthy since the \vae is not constrained to produce
\emph{exact} \ipt{}s, hence its outputs may contain artefacts.
The fact that our \encoder model~(remaining fixed throughout this
experiment) can reconstruct point clouds from such out-of-sample data
\emph{without} additional \mbox{regularisation} or retraining, shows
that the \ipt describes a stable and expressive latent space.
We observe fast \emph{training times}, with the \vae taking
approximately $\SI{15}{\minute}$, making the full training pipeline run
in less than \SI{1}{\hour}.
This is in stark contrast to models like LION~(\SI{550}{\hour}) or
SoftFlow~($\approx\SI{144}{\hour}$).
Finally \cref{tab:performance} shows that our \emph{inference times} on
commodity hardware are orders of magnitude faster than existing models.

\begin{framed}
	The \ipt yields a suitable representation for generating high-quality
	point clouds, while exhibiting inference times that are orders of
	magnitude faster than existing models. More expressive generative
	models are likely to improve generative quality~(at the expense of
	longer runtimes).
\end{framed}

\begin{table*}[tbp]
	\centering
	\sisetup{
		detect-all              = true,
retain-zero-uncertainty = true,
	}\caption{Evaluation metrics for point cloud generation tasks, cited from
		their respective papers.
We highlight the best result in \textbf{bold} and the second-best in \textit{italics}.
Our implementation of the EMD follows PVD~\citep{zhou2021pvd}.
Whenever available, we report coverage~(COV), minimum matching
		distance~(MMD), and \mbox{$1$-NNA}~($1$-nearest neighbour accuracy)
		for both EMD and CD.
	}\label{tab:shapenet_gen_full}
\let\b\bfseries
	\let\i\itshape
	\resizebox{\textwidth}{!}{\begin{tabular}{l
			S[table-format=1.2]@{\hspace{6pt}}S[table-format=1.2]
			S[table-format=2.2]@{\hspace{6pt}}S[table-format=2.2]
			S[table-format=2.2]@{\hspace{6pt}}S[table-format=2.2]
S[table-format=1.2]@{\hspace{6pt}}S[table-format=1.2]
			S[table-format=2.2]@{\hspace{6pt}}S[table-format=2.2]
			S[table-format=2.2]@{\hspace{6pt}}S[table-format=2.2]
S[table-format=1.2]@{\hspace{6pt}}S[table-format=1.2]
			S[table-format=2.2]@{\hspace{6pt}}S[table-format=2.2]
			S[table-format=2.2]@{\hspace{6pt}}S[table-format=2.2]
			}
			\toprule
			            & \multicolumn{6}{c}{\emph{Airplane}}
			            & \multicolumn{6}{c}{\emph{Chair}}
			            & \multicolumn{6}{c}{\emph{Car}}                                                                                                                                                                                                                   \\
			\midrule & \multicolumn{2}{c}{\small MMD\,($\downarrow$)}
			            & \multicolumn{2}{c}{\small COV\,($\uparrow$)}
			            & \multicolumn{2}{c}{\small 1-NNA\,($\downarrow$)}
			            & \multicolumn{2}{c}{\small MMD\,($\downarrow$)}
			            & \multicolumn{2}{c}{\small COV\,($\uparrow$)}
			            & \multicolumn{2}{c}{\small 1-NNA\,($\downarrow$)}
			            & \multicolumn{2}{c}{\small MMD\,($\downarrow$)}
			            & \multicolumn{2}{c}{\small COV\,($\uparrow$)}
			            & \multicolumn{2}{c}{\small 1-NNA\,($\downarrow$)}                                                                                                                                                                                                 \\
			\midrule
			Model
			            & {\small CD }
			            & {\small EMD }
			            & {\small CD }
			            & {\small EMD }
			            & {\small CD }
			            & {\small EMD }
			            & {\small CD }
			            & {\small EMD }
			            & {\small CD }
			            & {\small EMD }
			            & {\small CD }
			            & {\small EMD }
			            & {\small CD }
			            & {\small EMD }
			            & {\small CD }
			            & {\small EMD }
			            & {\small CD }
			            & {\small EMD }                                                                                                                                                                                                                                    \\
			\midrule
PointFlow   & 0.2243                                           & 0.3901    & \i 47.90 & 46.41    & 75.68    & 69.44     & 2.409    & 1.595    & 42.90    & 50.00    & 60.88    & 59.89    & 0.9010    & 0.8071    & 46.88     & 50.00    & 60.65    & 62.36    \\
			SoftFlow    & 0.2309                                           & 0.3745    & 46.91    & 47.90    & 70.92    & 69.44     & \i 2.528 & 1.682    & 41.39    & 47.43    & 59.95    & 63.51    & 1.187     & 0.8594    & 42.90     & 44.60    & 62.63    & 64.71    \\
			Shape-GF    & 2.703                                            & 0.6592    & 40.74    & 40.49    & 80.00    & 76.17     & 2.889    & 1.702    & 46.67    & 48.03    & 68.96    & 65.48    & 9.232     & 0.7558    & \i 49.43  & 50.28    & 63.20    & 56.53    \\
			SetVAE      & \i 0.200                                         & 0.367     & 43.70    & 48.40    & 75.31    & 77.65     & 2.545    & 1.585    & 46.83    & 44.26    & 58.76    & 61.48    & \i 0.882  & \i 0.733  & 49.15     & 46.59    & 59.66    & 61.48    \\
			PVD         & 0.2243                                           & 0.3803    & \b 48.88 & \b 52.09 & 73.82    & 64.81     & 2.622    & 1.556    & \b 49.84 & \i 50.60 & 56.26    & 53.32    & 1.077     & 0.7938    & 41.19     & \i 50.56 & \i 54.55 & \i 53.83 \\
			LION        & 0.219                                            & \i 0.372  & 47.16    & \i 49.63 & \i 67.41 & 61.23     & 2.640    & \i 1.55  & \i 48.94 & \b 52.11 & \b 53.70 & \i 52.34 & 0.913     & 0.752     & \b  50.00 & \b 56.53 & \b 53.41 & \b 51.14 \\
			PSF         &                                                  &           &          &          & 71.11    & \i  61.09 &          &          &          &          & 58.92    & 54.45    &           &           &           &          & 57.19    & 56.07    \\
			XCube       &                                                  &           &          &          & \b 52.85 & \b 49.75  &          &          &          &          & \i 53.99 & \b 48.60 &           &           &           &          & 57.96    & 54.43    \\\midrule
\vae (Ours) & \b 0.1901                                        & \b 0.3383 & 45.67    & 45.67    & 77.28    & 68.40     & \b 2.336 & \b 1.479 & 41.24    & 48.34    & 61.33    & 63.07    & \b 0.8513 & \b 0.7019 & 35.23     & 49.72    & 58.52    & 59.23    \\
			\bottomrule
		\end{tabular}}\end{table*}
 
\subsection{Point Cloud Downsampling}
\label{sec:Downsampling}

Motivated by the promising results in terms of generative performance,
we further investigate the capacity of our \encoder model to upsample
a downsampled point cloud.

\paragraph{Architecture and experimental setup.}
\begin{wraptable}{l}{0.40\linewidth}
	\vspace{-1\baselineskip}\centering
	\sisetup{
		detect-all              = true,
		table-format            = 2.2,
		detect-mode             = true,
		separate-uncertainty    = true,
		retain-zero-uncertainty = true,
		mode                    = text,
	}\caption{Reconstruction performance for consecutive down- and upsampling, with the best
		result shown in \textbf{bold}.
For reference purposes, the last row repeats our results without downsampling.
	}\label{tab:shapenet_reconstruction_downsample}
\let\b\bfseries
	\let\i\itshape
	\resizebox{\linewidth}{!}{\begin{tabular}{l
S[table-format=1.2]@{\hspace{6pt}}S[table-format=1.2]S[table-format=2.2]@{\hspace{6pt}}S[table-format=2.2]S[table-format=1.2]@{\hspace{6pt}}S[table-format=1.2]}
			\toprule
			                  & \multicolumn{2}{c}{\emph{Airplane}{\small\,($\downarrow$)}}
			                  & \multicolumn{2}{c}{\emph{Chair}{\small\,($\downarrow$)}}
			                  & \multicolumn{2}{c}{\emph{Car}{\small\,($\downarrow$)}}                                                          \\\midrule
			{\textsc{Method}} & {\small CD }
			                  & {\small EMD}
			                  & {\small CD }
			                  & {\small EMD}
			                  & {\small CD }
			                  & {\small EMD}                                                                                                    \\\midrule
			\downsampler      & \b 1.18                                                     & \b 2.58 & \b 11.70 & \b 11.42 & \b 6.42 & \b 5.81 \\
Uniform           & 3.10                                                        & 4.71    & 15.38    & 15.34    & 9.30    & 9.09    \\
			\midrule
			\encoder          & 1.03                                                        & 1.46    & 9.53     & 8.45     & 6.12    & 4.17    \\
\bottomrule
		\end{tabular}}\end{wraptable}
 We train a downsampling model with the same architecture as the
\encoder, resulting in a point cloud with $256$ points.
Here, the purpose of the loss term is to minimise the
discrepancy of the original point cloud to its downsampled version,
measured using the \ipt, which we normalise because of its dependency on
point cloud cardinality.
We then pass the \ipt of the downsampled point cloud to the
\encoder model to obtain an upsampled version, which, ideally, should be
close to the original point cloud.
Reporting both CD and EMD scores, we compare these results to a uniform
subsampling method.

\paragraph{Results.}
\cref{tab:shapenet_reconstruction_downsample} depicts numerical
scores~(cf.\ \cref{app:downsample} in the appendix for sample point
clouds).
Unlike uniform subsampling, our downsampling model preserves uniform
density of each object, showing that it has learned the underlying
shape.
We observe some loss in quality compared to reconstruction
without downsampling, which is to be expected. Despite potential compounding
errors, the loss in quality remains low.
Due to the similar setup, we can compare results with
\cref{tab:shapenet_reconstruction} and we observe that even when
downsampling, we \emph{still} perform on a par with the best model in
the reconstruction task.
Notably, the \encoder has not been retrained for this task, so the \ipt
of the subsampled point cloud is out-of-distribution.
This indicates that our model has captured the `true' underlying
shape characteristics.

\begin{framed}
	Our \encoder can downsample point clouds with a minimal loss of
	quality, demonstrating that the model is effective in encoding
	relevant shape properties.
\end{framed}

\subsection{Practical Inversion of the \ipt{}}
\label{sec:Backpropagation}

While theoretical invertibility~(see \cref{thm:iptinjective}) ensures that
the \ipt yields an \emph{expressive} summary, practical algorithms for
inversion of an \ipt remain an open question.
However, in our setting, making use of an approximation to the \ipt,
i.e.\ $\widehat{\mathrm{\ipt}}$, we may use backpropagation to invert it
\emph{without} requiring an \encoder.
The core observation is the differentiability of the
$\widehat{\mathrm{\ipt}}$ with respect to the point cloud coordinates,
and the fact that it constitutes a metric~(see \cref{lem:iptmetric}).
The main idea is to randomly initialise a point cloud, compute its
$\widehat{\mathrm{\ipt}}$, and subsequently calculate the loss between
said $\widehat{\mathrm{\ipt}}$ and the $\widehat{\mathrm{\ipt}}$ of the
target point cloud.
Differentiability ensures we can compute the derivative with respect to the 
point cloud coordinates and update them to reduce this loss. 
Applying backpropagation thus ultimately results in convergence to the
original point cloud. 
We show that
\begin{inparaenum}[(i)]
	\item \ipt{}s can be inverted in practice,
	\item a resolution of $128$ yields high-quality reconstructions, and
	\item the \ipt has \emph{stable gradients}~(which is not clear
	a priori, given that we are approximating an indicator function).
\end{inparaenum}
For this experiment, we apply backpropagation for $2000$ epochs with the Adam
optimiser using a learning rate of $0.5$, which is halved in epochs $50$,
$100$, $200$, and $1000$, respectively.
We repeat all experiments with three resolutions and for three choices
of scaling factor~$\lambda$ in \cref{eq:IPTIndicator}, which we set
either to be equal, half, or a quarter of selected resolution.
As a quality metric, we report the EMD and CD between the optimised
and target point cloud.

\paragraph{Results.}
\begin{wraptable}[9]{l}{0.33\linewidth}
	\centering
	\vspace{-1\baselineskip}\sisetup{
		detect-all              = true,
		table-format            = 2.2(1.2),
		detect-mode             = true,
		separate-uncertainty    = true,
		retain-zero-uncertainty = true,
		mode                    = text,
	}\let\b\bfseries
	\let\i\itshape
\caption{Inversion of the IPT with backpropagation leads to high quality 
		reconstruction.
	}\resizebox{\linewidth}{!}{\begin{tabular}{l
S[table-format=1.2(1.2)]S[table-format=1.2(1.2)]
}
			\toprule
			{\small\sc Dataset} & {\small CD ($\downarrow$)} & {\small EMD ($\downarrow$)} \\
			\midrule
			\emph{Airplane}     & 0.52 \pm 0.0               & 0.12 \pm 0.0                \\       
			\emph{Car}          & 2.04 \pm 0.0               & 0.39 \pm 0.0                \\
			\emph{Chair}        & 1.83 \pm 0.0               & 0.35 \pm 0.0                \\
			\bottomrule
		\end{tabular}
	}\label{tab:ShapeNet Render}
\end{wraptable}

 \cref{tab:ShapeNet Render} and \cref{fig:shapenetcore_rendered} in
the appendix depict the numerical results and show some reconstructed
samples, respectively.
Comparing the result with \cref{tab:shapenet_reconstruction}, we observe 
that backpropagation outperforms the \encoder model. 
\cref{tab:shapenet_render_full} in the appendix provides the full set of
results and an ablation with respect to the scale and resolution.
We observe that all approximations become gradually coarser as the
resolution is decreased.
As expected, we observe problems with vanishing gradients for
large values of~$\lambda$, since the sigmoid function approaches an
indicator function in this case.
This requires balancing $\lambda$ with respect to the resolution, and we
empirically observed that around a quarter of the \ipt resolution is
sufficient for high-quality results, which is how we pick the parameter
in practice.
It is important to note that while this method of inverting works for an \ipt
without any artifacts, it fails to accurately converge if the \ipt is only an
approximation of a true \ipt, as is the case with a \emph{generated} \ipt.
This also further motivates the use of a \emph{learned} method for reconstruction.

\begin{framed}
	Backpropagation through \ipt{}s directly is possible and outperforms the 
	\encoder in reconstruction quality.
However, this optimisation-based scheme is unable deal with noisy or generated \ipt{}s, 
	motivating the need of the \encoder.
\end{framed}

\begin{figure*}[tbp]
	\centering
	\includegraphics[width=0.1\textwidth]{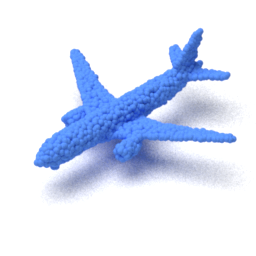}
	\includegraphics[width=0.1\textwidth]{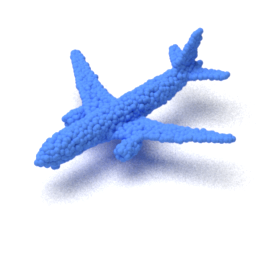}
	\includegraphics[width=0.1\textwidth]{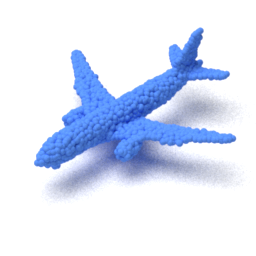}
	\includegraphics[width=0.1\textwidth]{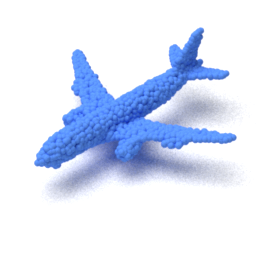}
	\includegraphics[width=0.1\textwidth]{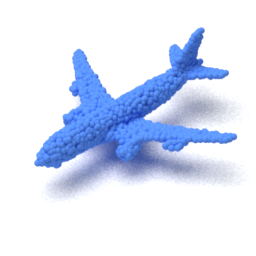}
	\includegraphics[width=0.1\textwidth]{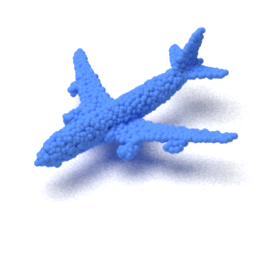}
	\includegraphics[width=0.1\textwidth]{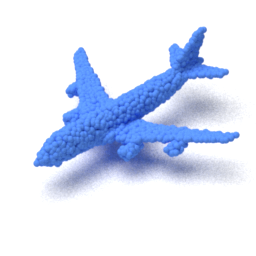}
	\includegraphics[width=0.1\textwidth]{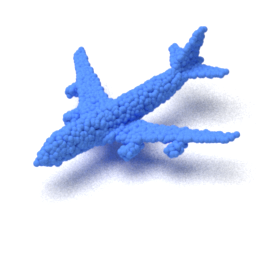}
	\includegraphics[width=0.1\textwidth]{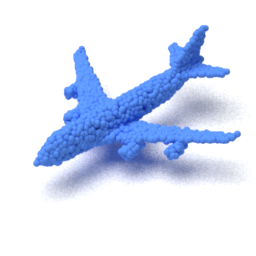}\newline
	\includegraphics[width=\textwidth]{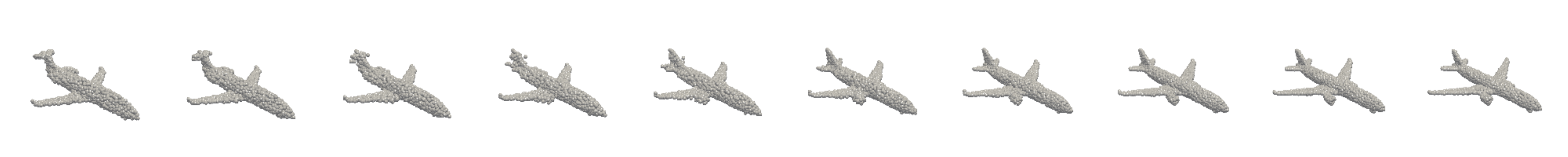}
	\caption{Linear interpolation between \ipt{}s results in a smooth interpolation in
		the point-cloud domain.
Given the \ipt of two airplanes, we linearly interpolate between the
		pixel values and pass each step through the \encoder to obtain
		a prediction of the intermediary point cloud.
Although the \encoder has not been specifically trained on such data,
		it is able to produce meaningful reconstructions, since it has learned
		the distribution of shapes.
We provide two examples of the interpolation process, one with high
    pairwise similarity~(top), the other with low pairwise
    similarity~(bottom).
	}\label{fig:shapenet_interpolation}
\end{figure*}
 
\subsection{Interpolation in Latent Space}
\label{sec:Interpolation}
As our final experiment, we consider the linear \emph{interpolation}
between the \ipt of two point clouds.
In the setting of infinite directions, \cref{thm:iptsurjective}
guarantees that we only encounter valid \ipt{}s along each
interpolation.
However, in practice, given our use of image-generative models, it is
not guaranteed that our input is an \ipt in the \emph{strict} sense.
Hence, it is crucial to understand the characteristic properties of the
\ipt in practice since not every sum of sigmoid functions is
guaranteed to correspond to a proper shape~(in other words, \ipt{}s are
not \emph{surjective}).
In our discretised representation, interpolating between \ipt{}s can be
achieved by interpolating per pixel.
To visually assess the quality of the latent space, we reconstruct each
intermediary \ipt during the interpolation using the \encoder.
\cref{fig:shapenet_interpolation} depicts the resulting point clouds; we
observe that the intermediary point clouds still remain plausible
reconstructions. Overall, this serves to highlight the utility of the
\ipt latent space.

\begin{framed}
	The \ipt permits linear interpolations, resulting in smooth
	interpolations between point clouds.
\end{framed}

\section{Discussion} \label{sec:Discussion}
We propose the \emph{Inner Product Transform}~(\ipt), which enables us
to represent point clouds using 2D images by evaluating inner products.
Next to showing that this representation has advantageous theoretical
properties, we develop an end-to-end-trainable pipeline for
\emph{point-cloud generation}.
Notably, we simplify the point-cloud generation process
by introducing an intermediate step, the generation of the \ipt
descriptor, for which one may use \emph{any} image-generation model.
Our method is
\begin{inparaenum}[(i)]
	\item exceptionally fast since it uses only inner products and other
	simple computational building blocks,
	\item well-grounded in theory~(it is a sufficient statistic on the
	space of point clouds), and,
	\item we observe that its \emph{reconstruction performance} and its
	\emph{generative capabilities} are, for the most part, on a par with
	more complex models.
\end{inparaenum}
This is due to the conceptual simplicity of the model and its provably
highly-stable latent space, which facilitates performing
out-of-distribution tasks \emph{without} the need for retraining or
additional regularisation.

\paragraph{Limitations.}
An unfortunate limitation arising from the disconnect between theory and
practice is that our injectivity results do not guarantee that one can
\emph{find} suitable directions, in particular given the complex
interplay between \emph{directions} and \emph{resolutions} of the \ipt.
We believe that additional guidance for selecting such parameters would
be helpful.
The current formulation of the \ipt is also \emph{not} invariant with
respect to rotations, which could be a desired property for some
applications~(or, at the very least, might make the method more robust
in the regime of smaller sample sizes). However, with increasing sample
size, we observe that \emph{data augmentation} enables \ipt-based models
to `learn' equivariance despite not being intrinsically invariant~(see
\cref{fig:MNIST rotated} in the appendix for preliminary results).
Moreover, if point clouds become high-dimensional, our
direction-sampling procedure becomes inefficient---this is not a problem
for the shape-generation tasks we are tackling in this paper, though.
A larger limitation in point-cloud generation involves the question of
evaluation metrics: In the absence of ground-truth information, it is
hard to choose the `best' model~(unlike point-cloud reconstruction,
where well-defined metrics exist).
In terms of other generative-model evaluation metrics like COV and \mbox{1-NNA},
our models exhibits only middling performance, ranking slightly better
in terms of COV than in terms of \mbox{1-NNA}. This is in stark contrast
to its excellent reconstruction performance and its strong MMD scores.
Taken together, this indicates that the $\widehat{\mathrm{\ipt}}$ is
capable of generating high-quality point clouds, but its generative
part, based on simple VAEs, is not sufficiently powerful to provide
\emph{diverse} representations.

\paragraph{Future work.}
Hence, future research concerning more complex---and expressive---model
architectures and their capabilities would be intriguing and potentially
lead to improved results in terms of diversity metrics, albeit at the
cost of computational efficiency.
We believe \emph{diffusion models} to be particularly suitable for
generating new \ipt{}s, given their proven track record in image
generation. Another possible extension of the \encoder could involve
a transformer-based architecture~\citep{Vaswani17a}, due to its high capacity for tasks
involving sets. This could be achieved by considering the \ipt as a bag
of tokens with associated directions.
Along those lines, the selected directions for computing the \ipt
would benefit from a proper \emph{positional encoding}, as opposed to
our implicit~(but simple) encoding; this would enable a model to better
capture dependencies between individual directions.
In addition, given the connection of the \ipt to more involved
geometrical-topological descriptors for graphs, meshes, or higher-order
complexes, a natural question is to what extent our inversion results
apply to such data.
This would further serve to contextualise the \ipt within the emerging
field of topological deep learning~\citep{papamarkou2024position} but
requires extensions to handle datasets without strong geometrical
signals~\citep{Ballester25a}, for instance.
Another feasible next step would be the extension to \emph{graph
generation tasks}, thus building a bridge between \emph{image
generation} and \emph{graph generation}. The caveats we raised about
generative metrics are all the more relevant in the context of graph
learning, though~\citep{Krimmel25a, OBray22a}.
Finally, while we have not yet explored any \emph{single-shot} or even
\emph{zero-shot} experiments, we believe that our work
may pave the path towards novel, efficient \emph{point cloud foundation models}.
 
\clearpage

\section*{Acknowledgments}
The authors are grateful for the productive discussions with the
reviewers and their useful feedback, which has significantly
strengthened our work. We are indebted to them and the area chair, who,
like us, believe in the merits of this paper. The authors are also
grateful for the thorough feedback received from Julius von Rohrscheidt.
This work has received funding from the Swiss State Secretariat for
Education, Research, and Innovation~(SERI).
 
\section*{Impact Statement}

This paper presents work whose goal is to advance the field of machine
learning, specifically the problem of high-quality point cloud~(or
shape) generation.
Given the computational efficiency of our method, we believe it to be
beneficial for `democratising' access to machine learning
algorithms.
There are many other potential societal consequences of point cloud
generation methods, none which we feel must be specifically highlighted
here.

\bibliographystyle{icml2025}
\bibliography{bibliography}

\begin{thebibliography}{44}
\providecommand{\natexlab}[1]{#1}
\providecommand{\url}[1]{\texttt{#1}}
\expandafter\ifx\csname urlstyle\endcsname\relax
  \providecommand{\doi}[1]{doi: #1}\else
  \providecommand{\doi}{doi: \begingroup \urlstyle{rm}\Url}\fi

\bibitem[Abramson et~al.(2024)Abramson, Adler, Dunger, Evans, Green, Pritzel,
  Ronneberger, Willmore, Ballard, Bambrick, Bodenstein, Evans, Hung, O'Neill,
  Reiman, Tunyasuvunakool, Wu, {\v{Z}}emgulyt{\.{e}}, Arvaniti, Beattie,
  Bertolli, Bridgland, Cherepanov, Congreve, Cowen-Rivers, Cowie, Figurnov,
  Fuchs, Gladman, Jain, Khan, Low, Perlin, Potapenko, Savy, Singh, Stecula,
  Thillaisundaram, Tong, Yakneen, Zhong, Zielinski, {\v{Z}}{\'i}dek, Bapst,
  Kohli, Jaderberg, Hassabis, and Jumper]{Abramson24a}
Abramson, J., Adler, J., Dunger, J., Evans, R., Green, T., Pritzel, A.,
  Ronneberger, O., Willmore, L., Ballard, A.~J., Bambrick, J., Bodenstein,
  S.~W., Evans, D.~A., Hung, C.-C., O'Neill, M., Reiman, D., Tunyasuvunakool,
  K., Wu, Z., {\v{Z}}emgulyt{\.{e}}, A., Arvaniti, E., Beattie, C., Bertolli,
  O., Bridgland, A., Cherepanov, A., Congreve, M., Cowen-Rivers, A.~I., Cowie,
  A., Figurnov, M., Fuchs, F.~B., Gladman, H., Jain, R., Khan, Y.~A., Low, C.
  M.~R., Perlin, K., Potapenko, A., Savy, P., Singh, S., Stecula, A.,
  Thillaisundaram, A., Tong, C., Yakneen, S., Zhong, E.~D., Zielinski, M.,
  {\v{Z}}{\'i}dek, A., Bapst, V., Kohli, P., Jaderberg, M., Hassabis, D., and
  Jumper, J.~M.
\newblock Accurate structure prediction of biomolecular interactions with {
  AlphaFold 3}.
\newblock \emph{Nature}, 630\penalty0 (8016):\penalty0 493--500, 2024.

\bibitem[Achlioptas et~al.(2018)Achlioptas, Diamanti, Mitliagkas, and
  Guibas]{achlioptas2018learning}
Achlioptas, P., Diamanti, O., Mitliagkas, I., and Guibas, L.
\newblock Learning representations and generative models for 3{D} point clouds.
\newblock In Dy, J. and Krause, A. (eds.), \emph{Proceedings of the 35th
  International Conference on Machine Learning}, volume~80 of \emph{Proceedings
  of Machine Learning Research}, pp.\  40--49. PMLR, 2018.

\bibitem[Amboage et~al.(2025)Amboage, Röell, Schnider, and Rieck]{Amboage25a}
Amboage, J., Röell, E., Schnider, P., and Rieck, B.
\newblock {LEAP}: Local {ECT}-based learnable positional encodings for graphs,
  2025.
\newblock URL \url{https://arxiv.org/abs/2510.00757}.

\bibitem[Am{\'e}zquita et~al.(2021)Am{\'e}zquita, Quigley, Ophelders, Landis,
  Koenig, Munch, and Chitwood]{amezquita2022measuring}
Am{\'e}zquita, E.~J., Quigley, M.~Y., Ophelders, T., Landis, J.~B., Koenig, D.,
  Munch, E., and Chitwood, D.~H.
\newblock Measuring hidden phenotype: {Q}uantifying the shape of barley seeds
  using the {E}uler characteristic transform.
\newblock \emph{in silico Plants}, 4\penalty0 (1):\penalty0 diab033, 12 2021.

\bibitem[Ballester et~al.(2025)Ballester, Röell, Schmid, Alain, Escalera,
  Casacuberta, and Rieck]{Ballester25a}
Ballester, R., Röell, E., Schmid, D.~B., Alain, M., Escalera, S., Casacuberta,
  C., and Rieck, B.
\newblock {MANTRA}: The manifold triangulations assemblage.
\newblock In \emph{International Conference on Learning Representations}, 2025.

\bibitem[Cai et~al.(2020)Cai, Yang, Averbuch-Elor, Hao, Belongie, Snavely, and
  Hariharan]{cai2020learning}
Cai, R., Yang, G., Averbuch-Elor, H., Hao, Z., Belongie, S., Snavely, N., and
  Hariharan, B.
\newblock Learning gradient fields for shape generation.
\newblock In Vedaldi, A., Bischof, H., Brox, T., and Frahm, J.-M. (eds.),
  \emph{Computer Vision -- ECCV 2020}, pp.\  364--381, Cham, Switzerland, 2020.
  Springer.

\bibitem[Cheng et~al.(2022)Cheng, Li, Liu, Sun, and Yang]{Cheng2022a}
Cheng, A.-C., Li, X., Liu, S., Sun, M., and Yang, M.-H.
\newblock Autoregressive {3D} shape generation via canonical mapping.
\newblock In Avidan, S., Brostow, G., Ciss{\'e}, M., Farinella, G.~M., and
  Hassner, T. (eds.), \emph{Computer Vision -- ECCV 2022}, pp.\  89--104, Cham,
  Switzerland, 2022. Springer.

\bibitem[Crawford et~al.(2020)Crawford, Monod, Chen, Mukherjee, and
  Rabad{\'a}n]{Crawford20a}
Crawford, L., Monod, A., Chen, A.~X., Mukherjee, S., and Rabad{\'a}n, R.
\newblock Predicting clinical outcomes in {G}lioblastoma: {A}n application of
  topological and functional data analysis.
\newblock \emph{Journal of the American Statistical Association}, 115\penalty0
  (531):\penalty0 1139--1150, 2020.

\bibitem[Curry et~al.(2022)Curry, Mukherjee, and Turner]{curry2022many}
Curry, J., Mukherjee, S., and Turner, K.
\newblock How many directions determine a shape and other sufficiency results
  for two topological transforms.
\newblock \emph{Transactions of the American Mathematical Society, Series B},
  9\penalty0 (32):\penalty0 1006--1043, 2022.

\bibitem[D{\l}otko(2024)]{Dlotko24a}
D{\l}otko, P.
\newblock On the shape that matters--topology and geometry in data science.
\newblock \emph{European Mathematical Society Magazine}, pp.\  5--13, 2024.

\bibitem[Fahim et~al.(2021)Fahim, Amin, and Zarif]{Fahim2021a}
Fahim, G., Amin, K., and Zarif, S.
\newblock Single-view {3D} reconstruction: {A} survey of deep learning methods.
\newblock \emph{Computers {\&} Graphics}, 94:\penalty0 164--190, 2021.

\bibitem[Fasy et~al.(2018)Fasy, Micka, Millman, Schenfisch, and
  Williams]{Fasy18a}
Fasy, B.~T., Micka, S., Millman, D.~L., Schenfisch, A., and Williams, L.
\newblock Challenges in reconstructing shapes from {E}uler characteristic
  curves, 2018.
\newblock URL \url{https://arxiv.org/abs/1811.11337}.

\bibitem[George et~al.(2025)George, Osborn, Munch, Ridgley~II, and
  Wang]{George25a}
George, J., Osborn, O.~L., Munch, E., Ridgley~II, M., and Wang, E.~X.
\newblock On the stability of the {E}uler characteristic transform for a
  perturbed embedding, 2025.
\newblock URL \url{https://arxiv.org/abs/2506.19991}.

\bibitem[Ghrist et~al.(2018)Ghrist, Levanger, and Mai]{Ghrist18a}
Ghrist, R., Levanger, R., and Mai, H.
\newblock Persistent homology and {E}uler integral transforms.
\newblock \emph{Journal of Applied and Computational Topology}, 2\penalty0
  (1):\penalty0 55--60, 2018.

\bibitem[Higgins et~al.(2017)Higgins, Matthey, Pal, Burgess, Glorot, Botvinick,
  Mohamed, and Lerchner]{Higgins2016betaVAELB}
Higgins, I., Matthey, L., Pal, A., Burgess, C., Glorot, X., Botvinick, M.,
  Mohamed, S., and Lerchner, A.
\newblock {$\beta$-VAE}: {L}earning basic visual concepts with a constrained
  variational framework.
\newblock In \emph{International Conference on Learning Representations}, 2017.

\bibitem[Kamb \& Ganguli(2025)Kamb and Ganguli]{Kamb25a}
Kamb, M. and Ganguli, S.
\newblock An analytic theory of creativity in convolutional diffusion models.
\newblock In Singh, A., Fazel, M., Hsu, D., Lacoste-Julien, S., Berkenkamp, F.,
  Maharaj, T., Wagstaff, K., and Zhu, J. (eds.), \emph{Proceedings of the 42nd
  International Conference on Machine Learning}, volume 267 of
  \emph{Proceedings of Machine Learning Research}, pp.\  28795--28831. PMLR,
  2025.

\bibitem[Kim et~al.(2020)Kim, Lee, Kang, Lee, and Kim]{kim2020softflow}
Kim, H., Lee, H., Kang, W.~H., Lee, J.~Y., and Kim, N.~S.
\newblock Soft{F}low: {P}robabilistic framework for normalizing flow on
  manifolds.
\newblock In Larochelle, H., Ranzato, M., Hadsell, R., Balcan, M., and Lin, H.
  (eds.), \emph{Advances in Neural Information Processing Systems}, volume~33,
  pp.\  16388--16397. Curran Associates, Inc., 2020.

\bibitem[Kim et~al.(2021)Kim, Yoo, Lee, and Hong]{kim2021setvae}
Kim, J., Yoo, J., Lee, J., and Hong, S.
\newblock {SetVAE}: {L}earning hierarchical composition for generative modeling
  of set-structured data.
\newblock In \emph{Proceedings of the IEEE/CVF Conference on Computer Vision
  and Pattern Recognition (CVPR)}, pp.\  15059--15068, 2021.

\bibitem[Krimmel et~al.(2025)Krimmel, Hartout, Borgwardt, and Chen]{Krimmel25a}
Krimmel, M., Hartout, P., Borgwardt, K., and Chen, D.
\newblock Polygraph discrepancy: {A} classifier-based metric for graph
  generation, 2025.
\newblock URL \url{https://arxiv.org/abs/2510.06122}.

\bibitem[Lee et~al.(2019)Lee, Lee, Kim, Kosiorek, Choi, and Teh]{Lee19a}
Lee, J., Lee, Y., Kim, J., Kosiorek, A., Choi, S., and Teh, Y.~W.
\newblock Set transformer: {A} framework for attention-based
  permutation-invariant neural networks.
\newblock In Chaudhuri, K. and Salakhutdinov, R. (eds.), \emph{Proceedings of
  the 36th International Conference on Machine Learning}, volume~97 of
  \emph{Proceedings of Machine Learning Research}, pp.\  3744--3753. PMLR,
  2019.

\bibitem[Liu et~al.(2019)Liu, Tang, Lin, and Han]{liu2019point}
Liu, Z., Tang, H., Lin, Y., and Han, S.
\newblock {Point-Voxel CNN} for efficient {3D} deep learning.
\newblock In Wallach, H., Larochelle, H., Beygelzimer, A., d~\textquotesingle
  Alch\'{e}-Buc, F., Fox, E., and Garnett, R. (eds.), \emph{Advances in Neural
  Information Processing Systems}, volume~32. Curran Associates, Inc., 2019.

\bibitem[Marsh et~al.(2024)Marsh, Zhou, Qin, Lu, Byrne, and
  Harrington]{Marsh24a}
Marsh, L., Zhou, F.~Y., Qin, X., Lu, X., Byrne, H.~M., and Harrington, H.~A.
\newblock Detecting temporal shape changes with the {E}uler characteristic
  transform.
\newblock \emph{Transactions of Mathematics and its Applications}, 8, 2024.

\bibitem[M\'{e}moli \& Sapiro(2004)M\'{e}moli and Sapiro]{memoli2004comparing}
M\'{e}moli, F. and Sapiro, G.
\newblock Comparing point clouds.
\newblock In \emph{Proceedings of the Eurographics/ACM SIGGRAPH Symposium on
  Geometry Processing}, pp.\  32--40, 2004.

\bibitem[Munch(2025)]{Munch23a}
Munch, E.
\newblock An invitation to the {E}uler characteristic transform.
\newblock \emph{The American Mathematical Monthly}, 132\penalty0 (1):\penalty0
  15--25, 2025.

\bibitem[Nadimpalli et~al.(2023)Nadimpalli, Chattopadhyay, and
  Rieck]{nadimpalli2023euler}
Nadimpalli, K.~V., Chattopadhyay, A., and Rieck, B.
\newblock {E}uler characteristic transform based topological loss for
  reconstructing {3D} images from single {2D} slices.
\newblock In \emph{Proceedings of the IEEE/CVF Conference on Computer Vision
  and Pattern Recognition Workshops~(CVPRW)}, pp.\  571--579, 2023.

\bibitem[O'Bray et~al.(2022)O'Bray, Horn, Rieck, and Borgwardt]{OBray22a}
O'Bray, L., Horn, M., Rieck, B., and Borgwardt, K.
\newblock Evaluation metrics for graph generative models: Problems, pitfalls,
  and practical solutions.
\newblock In \emph{International Conference on Learning Representations}, 2022.

\bibitem[\"Ozyeşil et~al.(2017)\"Ozyeşil, Voroninski, Basri, and
  Singer]{Oezyesil17a}
\"Ozyeşil, O., Voroninski, V., Basri, R., and Singer, A.
\newblock A survey of structure from motion.
\newblock \emph{Acta Numerica}, 26:\penalty0 305--364, 2017.

\bibitem[Papamarkou et~al.(2024)Papamarkou, Birdal, Bronstein, Carlsson, Curry,
  Gao, Hajij, Kwitt, Li{\`o}, Lorenzo, Maroulas, Miolane, Nasrin, Ramamurthy,
  Rieck, Scardapane, Schaub, Veli{\v{c}}kovi{\'c}, Wang, Wang, Wei, and
  Zamzmi]{papamarkou2024position}
Papamarkou, T., Birdal, T., Bronstein, M., Carlsson, G., Curry, J., Gao, Y.,
  Hajij, M., Kwitt, R., Li{\`o}, P., Lorenzo, P.~D., Maroulas, V., Miolane, N.,
  Nasrin, F., Ramamurthy, K.~N., Rieck, B., Scardapane, S., Schaub, M.~T.,
  Veli{\v{c}}kovi{\'c}, P., Wang, B., Wang, Y., Wei, G.-W., and Zamzmi, G.
\newblock Position: {T}opological deep learning is the new frontier for
  relational learning.
\newblock In Salakhutdinov, R., Kolter, Z., Heller, K., Weller, A., Oliver, N.,
  Scarlett, J., and Berkenkamp, F. (eds.), \emph{Proceedings of the 41st
  International Conference on Machine Learning}, number 235 in Proceedings of
  Machine Learning Research, pp.\  39529--39555. PMLR, 2024.

\bibitem[Qi et~al.(2017)Qi, Yi, Su, and Guibas]{Qi2017}
Qi, C.~R., Yi, L., Su, H., and Guibas, L.~J.
\newblock {PointNet++}: {D}eep hierarchical feature learning on point sets in a
  metric space.
\newblock In Guyon, I., Luxburg, U.~V., Bengio, S., Wallach, H., Fergus, R.,
  Vishwanathan, S., and Garnett, R. (eds.), \emph{Advances in Neural
  Information Processing Systems}, volume~30. Curran Associates, Inc., 2017.

\bibitem[Ren et~al.(2024)Ren, Huang, Zeng, Museth, Fidler, and
  Williams]{ren2024xcube}
Ren, X., Huang, J., Zeng, X., Museth, K., Fidler, S., and Williams, F.
\newblock Xcube: Large-scale 3d generative modeling using sparse voxel
  hierarchies.
\newblock In \emph{Proceedings of the IEEE/CVF Conference on Computer Vision
  and Pattern Recognition (CVPR)}, pp.\  4209--4219, June 2024.

\bibitem[Rieck(2025)]{Rieck25a}
Rieck, B.
\newblock Topology meets machine learning: {A}n introduction using the {E}uler
  characteristic transform.
\newblock \emph{Notices of the American Mathematical Society}, 72\penalty0
  (7):\penalty0 719--727, 2025.

\bibitem[R{\"o}ell \& Rieck(2024)R{\"o}ell and Rieck]{Roell24a}
R{\"o}ell, E. and Rieck, B.
\newblock Differentiable {E}uler characteristic transforms for shape
  classification.
\newblock In \emph{International Conference on Learning Representations}, 2024.

\bibitem[Tancik et~al.(2020)Tancik, Srinivasan, Mildenhall, Fridovich-Keil,
  Raghavan, Singhal, Ramamoorthi, Barron, and Ng]{Tancik20a}
Tancik, M., Srinivasan, P., Mildenhall, B., Fridovich-Keil, S., Raghavan, N.,
  Singhal, U., Ramamoorthi, R., Barron, J., and Ng, R.
\newblock Fourier features let networks learn high frequency functions in low
  dimensional domains.
\newblock In Larochelle, H., Ranzato, M., Hadsell, R., Balcan, M., and Lin, H.
  (eds.), \emph{Advances in Neural Information Processing Systems}, volume~33,
  pp.\  7537--7547. Curran Associates, Inc., 2020.

\bibitem[Toscano-Duran et~al.(2026)Toscano-Duran, Rottach, and
  Rieck]{Toscano25a}
Toscano-Duran, V., Rottach, F., and Rieck, B.
\newblock Molecular machine learning using {E}uler characteristic transforms.
\newblock In L{\'o}pez~Fern{\'a}ndez, A., Rodr{\'i}guez-Gonz{\'a}lez, A.,
  Leir{\'o}s-Rodr{\'i}guez, R., Mata~Miquel, C., and Gonz{\'a}lez~Su{\'a}rez,
  V.~M. (eds.), \emph{Artificial Intelligence in Biomedicine}, pp.\  391--405,
  Cham, Switzerland, 2026. Springer.

\bibitem[Turner et~al.(2014)Turner, Mukherjee, and Boyer]{Turner14a}
Turner, K., Mukherjee, S., and Boyer, D.~M.
\newblock Persistent homology transform for modeling shapes and surfaces.
\newblock \emph{Information and Inference: A Journal of the IMA}, 3\penalty0
  (4):\penalty0 310--344, 12 2014.

\bibitem[Vaswani et~al.(2017)Vaswani, Shazeer, Parmar, Uszkoreit, Jones, Gomez,
  Kaiser, and Polosukhin]{Vaswani17a}
Vaswani, A., Shazeer, N., Parmar, N., Uszkoreit, J., Jones, L., Gomez, A.~N.,
  Kaiser, L.~u., and Polosukhin, I.
\newblock Attention is all you need.
\newblock In Guyon, I., Luxburg, U.~V., Bengio, S., Wallach, H., Fergus, R.,
  Vishwanathan, S., and Garnett, R. (eds.), \emph{Advances in Neural
  Information Processing Systems}, volume~30. Curran Associates, Inc., 2017.

\bibitem[von Rohrscheidt \& Rieck(2025)von Rohrscheidt and
  Rieck]{vonRohrscheidt25a}
von Rohrscheidt, J. and Rieck, B.
\newblock {Diss-l-ECT}: Dissecting graph data with local {E}uler characteristic
  transforms.
\newblock In Singh, A., Fazel, M., Hsu, D., Lacoste-Julien, S., Berkenkamp, F.,
  Maharaj, T., Wagstaff, K., and Zhu, J. (eds.), \emph{Proceedings of the 42nd
  International Conference on Machine Learning}, volume 267 of
  \emph{Proceedings of Machine Learning Research}, pp.\  61790--61809. PMLR,
  2025.

\bibitem[Wu et~al.(2023)Wu, Wang, Gong, Liu, Xiong, Ranjan, Krishnamoorthi,
  Chandra, and Liu]{wu2023fast}
Wu, L., Wang, D., Gong, C., Liu, X., Xiong, Y., Ranjan, R., Krishnamoorthi, R.,
  Chandra, V., and Liu, Q.
\newblock Fast point cloud generation with straight flows.
\newblock In \emph{Proceedings of the IEEE/CVF Conference on Computer Vision
  and Pattern Recognition (CVPR)}, pp.\  9445--9454, 2023.

\bibitem[Xu et~al.(2023)Xu, Mu, and Yang]{Xu23a}
Xu, Q.-C., Mu, T.-J., and Yang, Y.-L.
\newblock A survey of deep learning-based {3D} shape generation.
\newblock \emph{Computational Visual Media}, 9\penalty0 (3):\penalty0 407--442,
  2023.

\bibitem[Yang et~al.(2019)Yang, Huang, Hao, Liu, Belongie, and
  Hariharan]{pointflow}
Yang, G., Huang, X., Hao, Z., Liu, M.-Y., Belongie, S., and Hariharan, B.
\newblock {P}oint{F}low: {3D} point cloud generation with continuous
  normalizing flows.
\newblock In \emph{IEEE/CVF International Conference on Computer
  Vision~(ICCV)}, pp.\  4540--4549, 2019.

\bibitem[Zaheer et~al.(2017)Zaheer, Kottur, Ravanbakhsh, Poczos, Salakhutdinov,
  and Smola]{zaheer2017deep}
Zaheer, M., Kottur, S., Ravanbakhsh, S., Poczos, B., Salakhutdinov, R.~R., and
  Smola, A.~J.
\newblock Deep sets.
\newblock In Guyon, I., Luxburg, U.~V., Bengio, S., Wallach, H., Fergus, R.,
  Vishwanathan, S., and Garnett, R. (eds.), \emph{Advances in Neural
  Information Processing Systems}, volume~30. Curran Associates, Inc., 2017.

\bibitem[Zeng et~al.(2022)Zeng, Vahdat, Williams, Gojcic, Litany, Fidler, and
  Kreis]{Vahdat2022a}
Zeng, X., Vahdat, A., Williams, F., Gojcic, Z., Litany, O., Fidler, S., and
  Kreis, K.
\newblock {LION}: {L}atent point diffusion models for {3D} shape generation.
\newblock In Koyejo, S., Mohamed, S., Agarwal, A., Belgrave, D., Cho, K., and
  Oh, A. (eds.), \emph{Advances in Neural Information Processing Systems},
  volume~35, pp.\  10021--10039. Curran Associates, Inc., 2022.

\bibitem[Zhang et~al.(2018)Zhang, Zhang, Zhang, Tenenbaum, Freeman, and
  Wu]{Zhang18a}
Zhang, X., Zhang, Z., Zhang, C., Tenenbaum, J., Freeman, B., and Wu, J.
\newblock Learning to reconstruct shapes from unseen classes.
\newblock In Bengio, S., Wallach, H., Larochelle, H., Grauman, K.,
  Cesa-Bianchi, N., and Garnett, R. (eds.), \emph{Advances in Neural
  Information Processing Systems}, volume~31. Curran Associates, Inc., 2018.

\bibitem[Zhou et~al.(2021)Zhou, Du, and Wu]{zhou2021pvd}
Zhou, L., Du, Y., and Wu, J.
\newblock {3D} shape generation and completion through {P}oint-{V}oxel
  {D}iffusion.
\newblock In \emph{Proceedings of the IEEE/CVF International Conference on
  Computer Vision (ICCV)}, pp.\  5826--5835, 2021.

\end{thebibliography}

\clearpage

\clearpage

\appendix

\counterwithin*{figure}{part}
\stepcounter{part}
\renewcommand{\thefigure}{S.\arabic{figure}}

\counterwithin*{table}{part}
\stepcounter{part}
\renewcommand{\thetable}{S.\arabic{table}}

\crefalias{section}{appendix}
\crefalias{subsection}{appendix}
\crefalias{subsubsection}{appendix}

\startcontents
\printcontents{}{1}{{\vskip10pt\hrule
			\large\textbf{Appendix}\vskip3pt\hrule\vskip5pt}
}
\clearpage

\section{Changelog}
\begin{figure}
	\center
  \includegraphics[width=\textwidth]{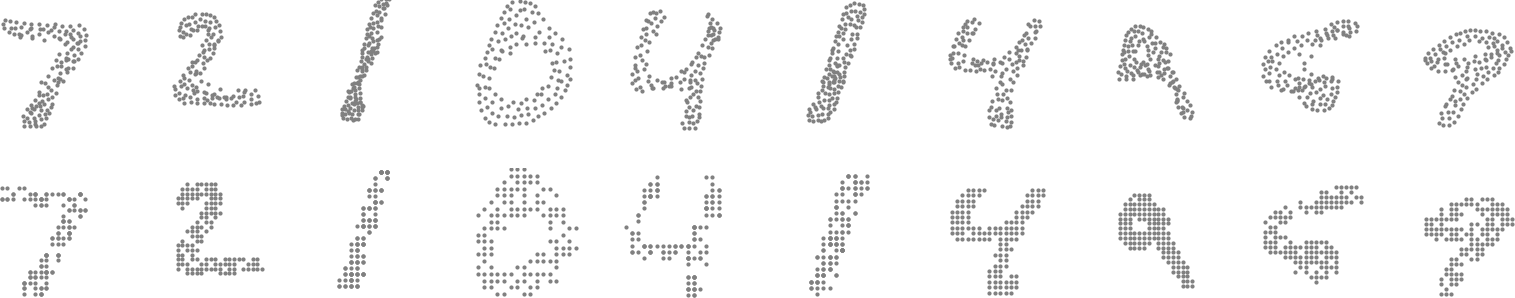}\\[1em]
	\includegraphics[width=\textwidth]{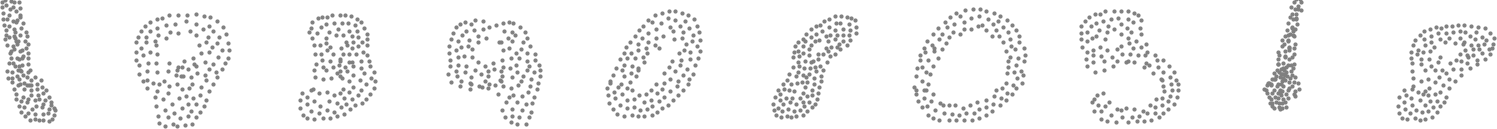}
	\caption{To ensure our pipeline can capture \emph{multi-class distributions},
		we train the \encoder and \vae on the MNIST dataset.
The top row depicts the reconstructed point clouds with the ground truth in the
		middle row.
The last row depicts point clouds generated with the \vae.
A latent vector is sampled from the latent distribution and decoded into an
		\ipt.
The novel \ipt{} is subsequently converted to a point cloud with the \encoder.
Using our evaluation we find the $1$-NNA CD and EMD scores to be
    $70.31$ and $62.50$, respectively. This indicates that our pipeline
    can handle multi-class data distributions.
	}
	\label{fig:MNIST generated}
\end{figure}

\begin{figure}
	\center
	\includegraphics[width=\textwidth]{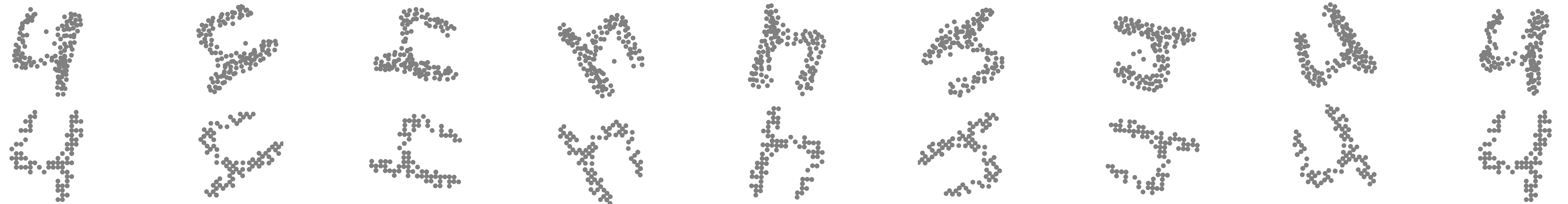}
	\caption{Recent work showed that \emph{equivariance} with respect to certain
		operations like rotations can also be achieved through data
		augmentation, thus obviating the need for more complex
		architectures~\citep{Abramson24a}.
To assess the capabilities of the \encoder in this context, we
		follow \citet{Qi2017} and use a point cloud version the MNIST dataset of
		handwritten digits.
During training, we apply a random rotation to each point cloud and then
		compute the \ipt{}, thus permitting the model to learn an equivariant
		representation of the data.
As the figure shows, data augmentation is \emph{sufficient} to encode
		rotations, resulting in an equivariant model without having to
		specifically add equivariance as a separate inductive bias.
}
	\label{fig:MNIST rotated}
\end{figure}

\begin{figure}[h]
	\center
	\includegraphics[width=\textwidth]{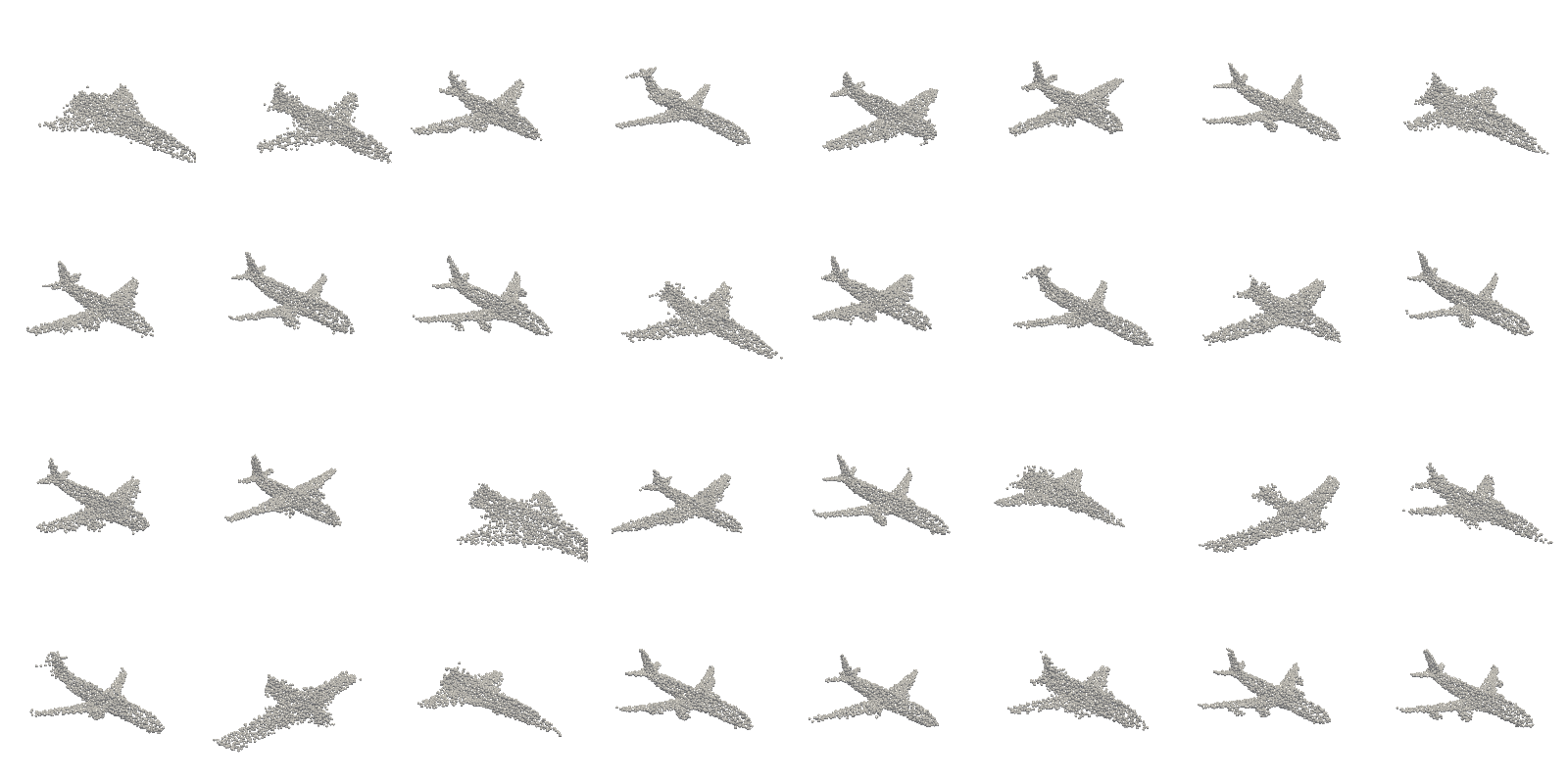}
	\caption{Preliminary results for generating samples of the \emph{airplane}
    class using a latent diffusion model.
We train a VAE on the set of \ipt{}s of airplanes, followed by
    training a diffusion model on the latent embeddings of the \ipt{}s
    to generate new \ipt{}s.
Again, we use the \encoder to map a generated \ipt into a point
    cloud. This demonstrates the flexibility and generality of our
    model, which we look forward to exploiting better in future work.
	}
	\label{fig:Latent Diffusion}
\end{figure}

We provide a brief list of changes that arose during the review process,
commenting shortly on how we implemented the feedback received by the
reviewers.
\begin{compactitem}
	\item Changes in the \emph{main text}:
	\begin{compactitem}
		\item We fixed the typo concerning 1D and 2D convolutions in \cref{sec:Methods} and \cref{sec:Experiments}.
		\item We updated the constraints \cref{thm:iptsurjective} to enforce
		$q$ to be non-zero.
		\item We fixed the typo concerning the training times in \cref{sec:Experiments}.
		\item We clarified all architectures in \cref{sec:Methods} and now discuss transformer-based models in \cref{sec:Discussion}.
		\item We clarified the ordering of directions in \cref{sec:Methods} and comment on improvements~(via \emph{positional encodings}) in \cref{sec:Discussion}.
    \item We added details about our hardware in \cref{sec:Experiments}. 
    \item We updated \cref{sec:Backpropagation} to better reflect the
      fact that we provide the first \emph{practical} method for inverting an
      \ipt.
	\end{compactitem}
\item Changes in the \emph{appendix}:
	\begin{compactitem}
		\item We added a new experiment on reconstructing point cloud from \emph{partial views} in \cref{tab:shapenet_completion}.
    \item We demonstrate \emph{out-of-distribution experiments} in \cref{tab:Shapenet Cross Reconstruction}. 
		\item We add a new ablation study with respect to the number of directions in \cref{tab:shapenet_num_dirs}. 
		\item We provided new results on downsampling in \cref{tab:shapenet_points_ablation}.
		\item	We provide new results on reconstructing large point clouds in \cref{tab:shapenet_points_ablation}~(bottom row).
    \item We added new results for \emph{multi-class generation} on the MNIST and ShapeNet13 datasets in \cref{fig:MNIST generated} and \cref{tab:shapenet_multiclass}.
		\item We provide a new experiment on \emph{learning} equivariance via \emph{data augmentation} in \cref{fig:MNIST rotated}.
		\item We wrote preliminary code for using latent diffusion models and added a first visualisation of the results in \Cref{fig:Latent Diffusion}.
	\end{compactitem}
\item We prefer to address the following aspects in \emph{future work}, rather than adding an inadequate discussion to our work:
  \begin{compactitem}
    \item A discussion on how to use spherical harmonics for the
      \ipt{}~(we believe that this could be essential for a completely
      new approach for calculating \ipt{}, beyond the scope of this
      paper).
    \item A full comparison with models like DeepSDF~(we lacked compute
      to finish this).
    \item A rate-distortion analysis~(we rewrote the compression aspects
      at the suggestion of the reviewers and we believe that such an
      analysis should be better contextualised in a follow-up work).
    \item A report of FLOPs~(we were unable to collect all this data for
      the larger baseline models in time and found some reporting
      inconsistencies).
  \end{compactitem}
\end{compactitem}
We very much appreciate the feedback by reviewers, which helped us
substantially improve our work.

\newpage

\section{Overview of the IPT}
\label{app:Overview of the IPT}
\begin{figure}[!ht]
	\centering
	\begin{subfigure}{.3\textwidth}
		\includegraphics[width=\linewidth]{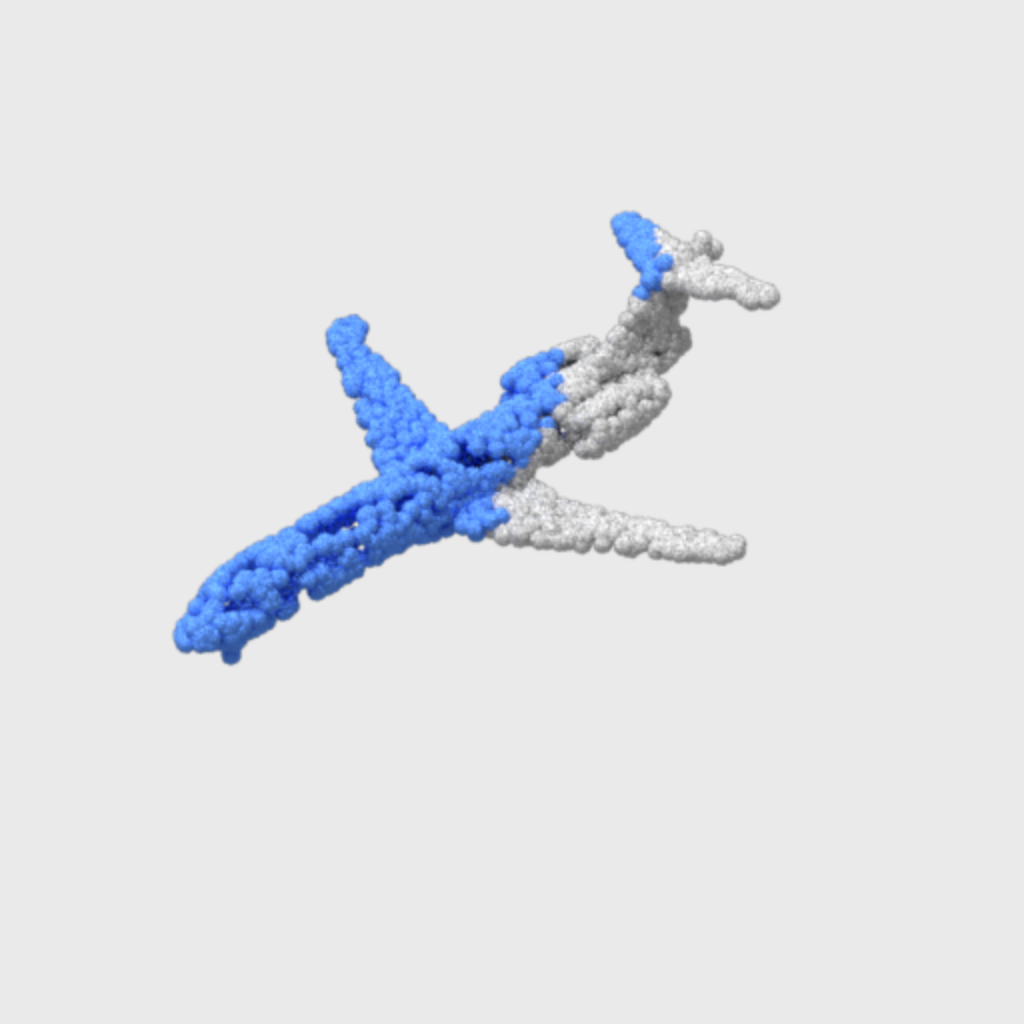}
		\subcaption{\phantom{x}}
		\label{sfig:Filtration}
	\end{subfigure}\hfill
	\begin{subfigure}{.3\textwidth}
		\includegraphics[width=\linewidth]{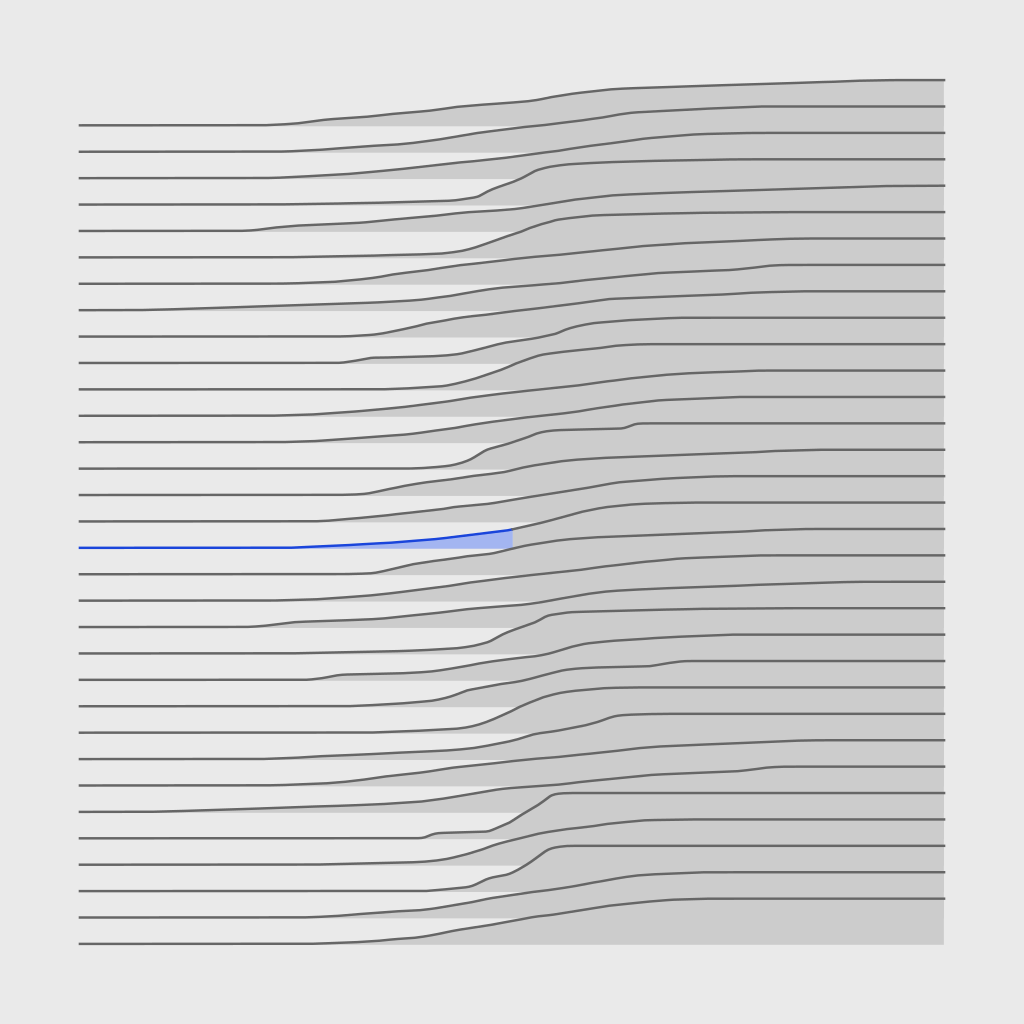}
		\subcaption{\phantom{x}}
		\label{sfig:Curves}
	\end{subfigure}\hfill
	\begin{subfigure}{.3\textwidth}
		\includegraphics[width=\linewidth]{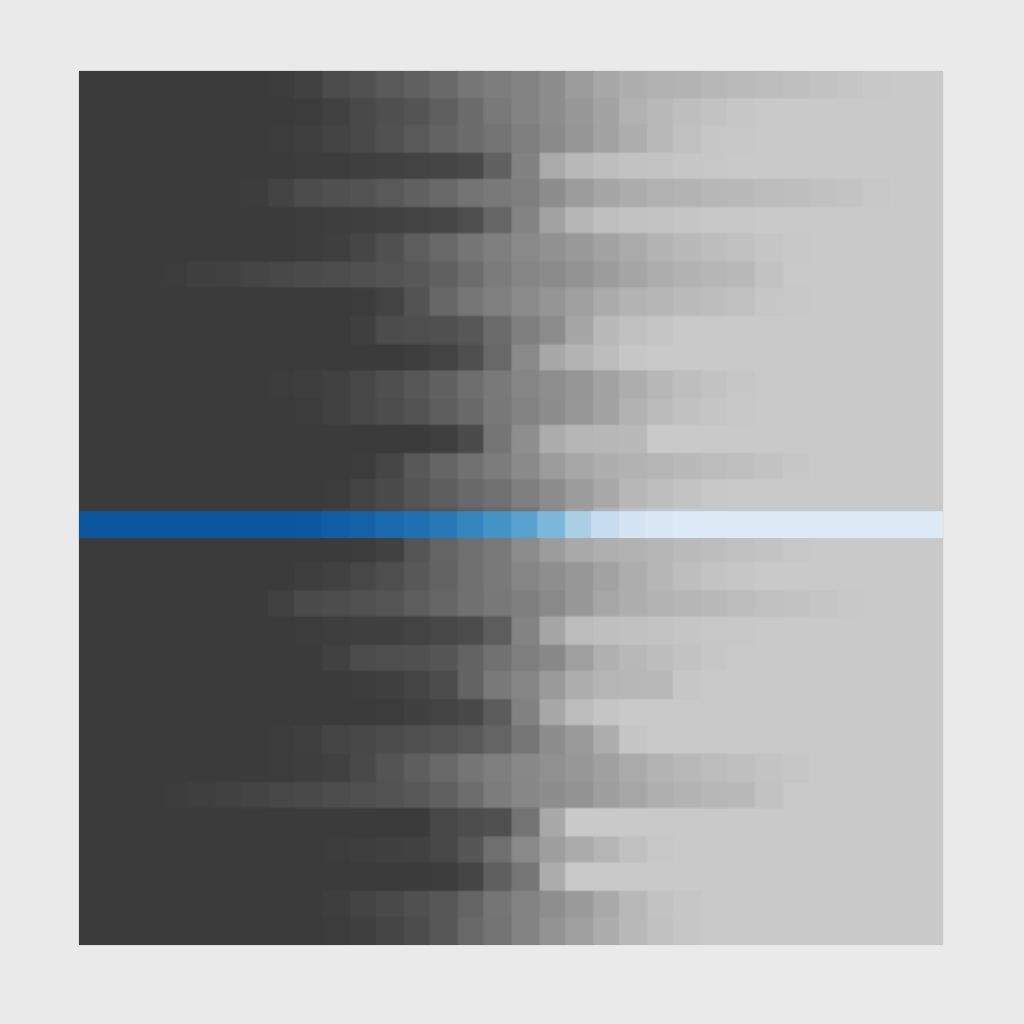}
		\subcaption{\phantom{x}}
		\label{sfig:Image}
	\end{subfigure}\hfill
\caption{An overview of the \ipt calculation for $32$ directions and
		resolution of $32$.
For a given direction vector $\xi$, we filter the point cloud with
		hyperplanes.
The partially filtered point cloud is shown in
		\subref{sfig:Filtration} and points included in the filtration are
		coloured {\color{airlanebleu}{blue}}.
For each of the $32$ directions, sampled uniformly from the sphere, the
		respective curves along each direction are shown in \subref{sfig:Curves}
		and the partially-completed curve from \subref{sfig:Filtration} is
		highlighted.
Note that neighbouring curves are \emph{not} necessarily related, requiring
		us to treat each curve as its own signal.
Each of the $32$ curves is discretised in $32$ steps and stacked to
		form an \emph{image} representation of the point cloud~\subref{sfig:Image} of
		size $32\times 32$.
The row corresponding to the full curve from \subref{sfig:Filtration} and
		each of the rows corresponds to the curve in the same index in \subref{sfig:Curves}.
For complex geometries such as the \shapenet data, we empirically observe
    that \emph{at least} $128$ directions are required, using a resolution of $128$.
	}\label{fig:ipt_overview}
\end{figure}
 \newpage

\section{Additional Downsampling Figures}
\label{app:Additional Figures}
We provide a visualisation of
\cref{tab:shapenet_reconstruction_downsample}
for the comparison of downsampling with the \downsampler~(left) and
uniform subsampling~(right) for all three classes.
\begin{figure}[!ht]
	\centering
\begin{subfigure}{.47\textwidth}
		\begin{subfigure}{.3\linewidth}
\includegraphics[width=\linewidth]{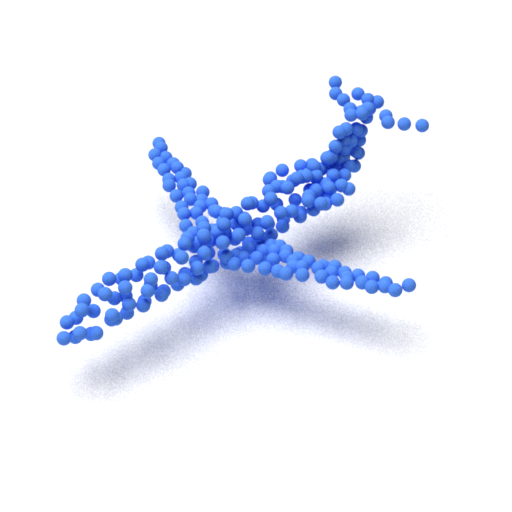}
			\includegraphics[width=\linewidth]{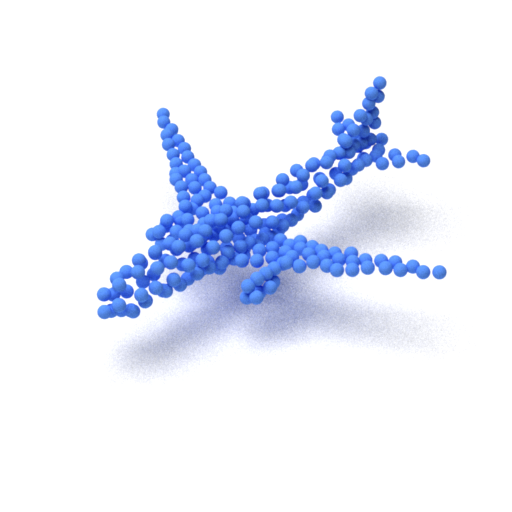}
			\includegraphics[width=\linewidth]{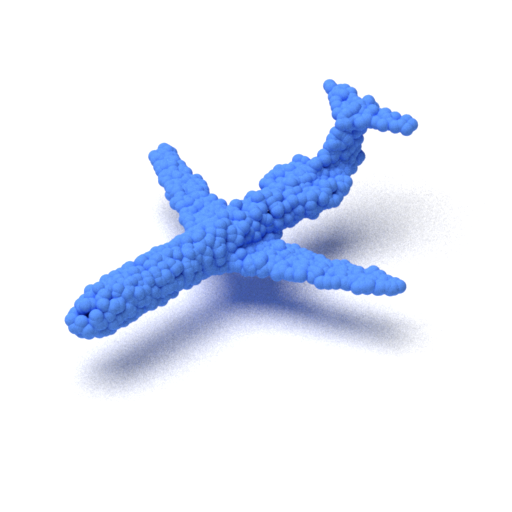}
			\includegraphics[width=\linewidth]{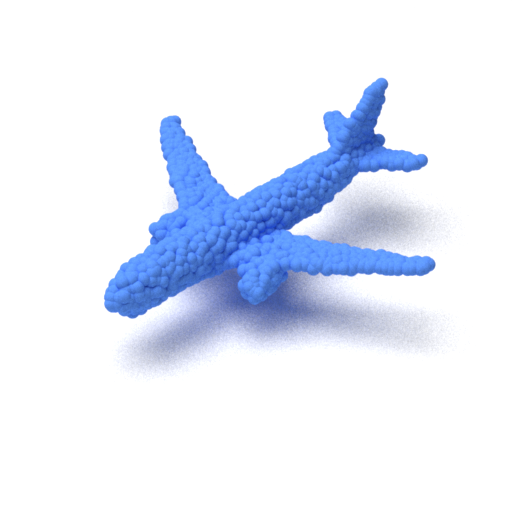}
		\end{subfigure}
		\begin{subfigure}{.3\linewidth}
\includegraphics[width=\linewidth]{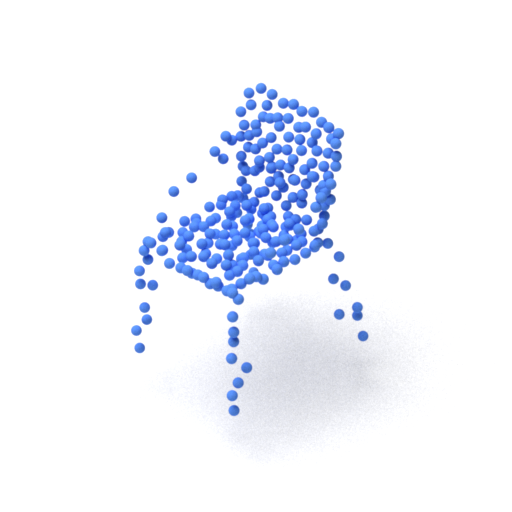}
			\includegraphics[width=\linewidth]{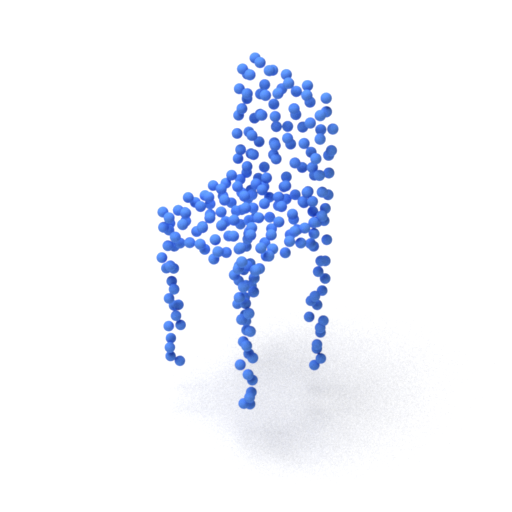}
			\includegraphics[width=\linewidth]{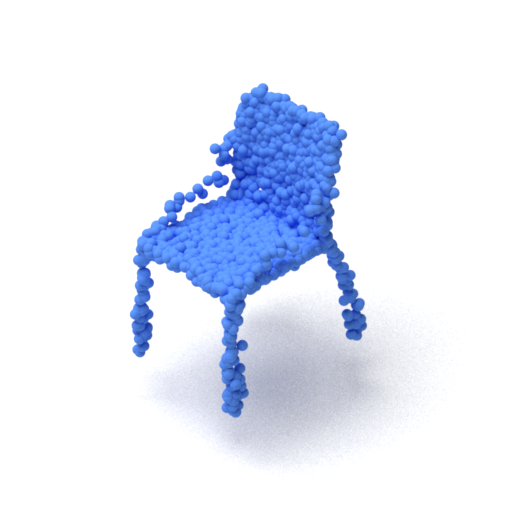}
			\includegraphics[width=\linewidth]{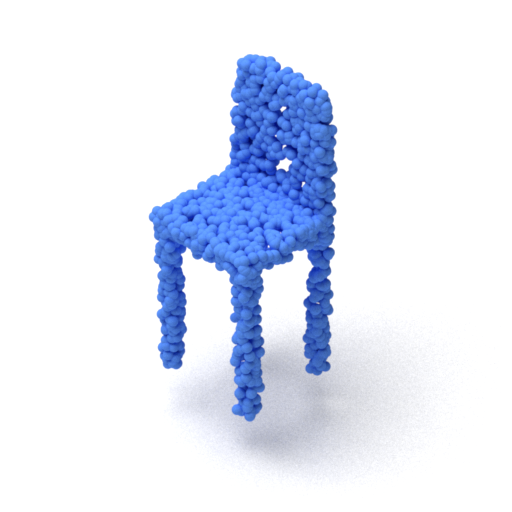}
		\end{subfigure}
		\begin{subfigure}{.3\linewidth}
\includegraphics[width=\linewidth]{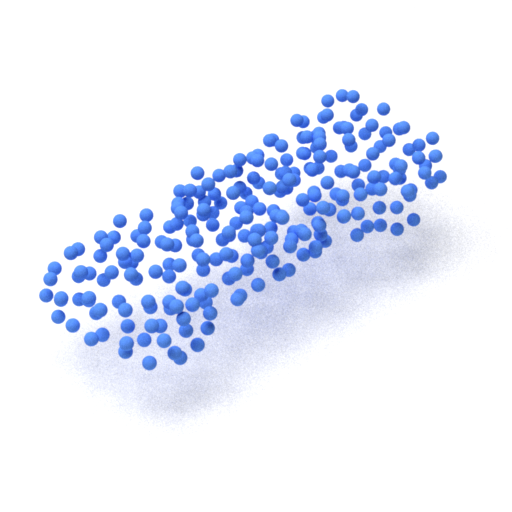}
			\includegraphics[width=\linewidth]{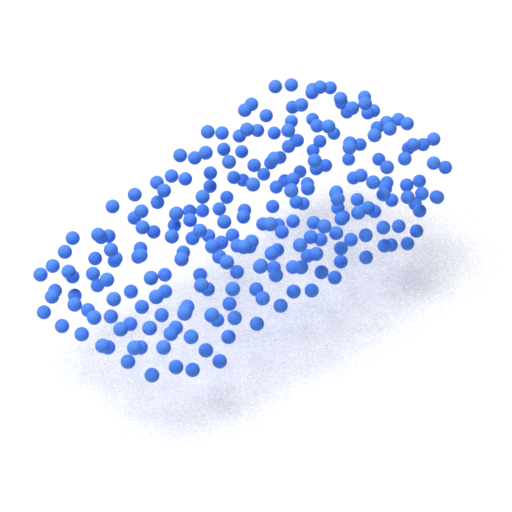}
			\includegraphics[width=\linewidth]{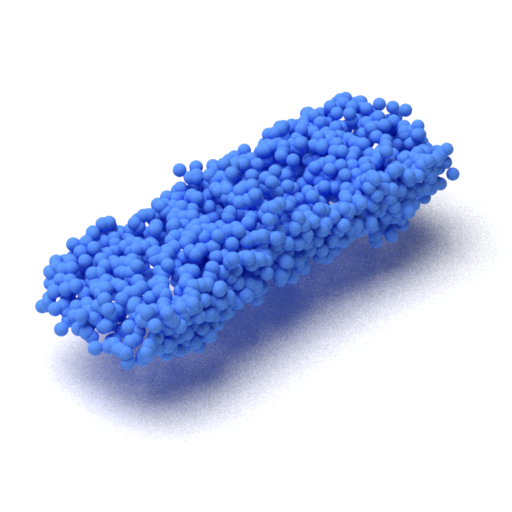}
			\includegraphics[width=\linewidth]{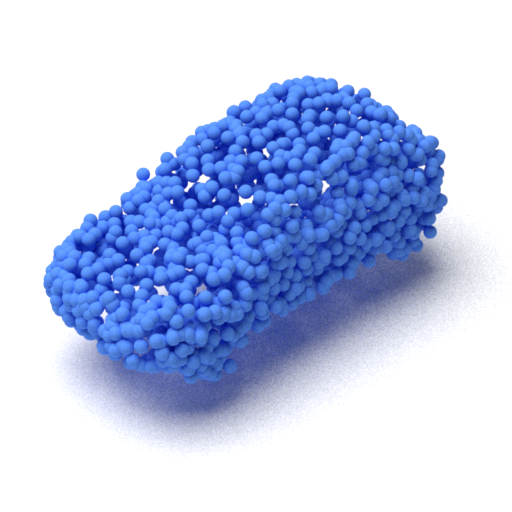}
		\end{subfigure}
		\subcaption{Downsampled point-clouds with the \downsampler~(top) and upsampled
			with the \encoder~(bottom).
		}\label{app:downsample_downsampler_appendix}
	\end{subfigure}\hfill
\begin{subfigure}{.47\textwidth}
		\begin{subfigure}{.3\textwidth}
\includegraphics[width=\linewidth]{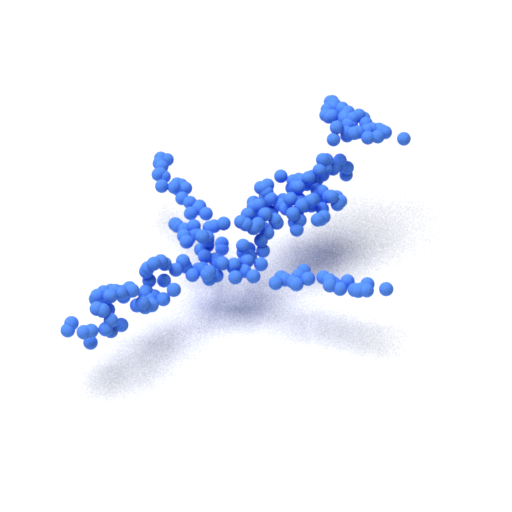}
			\includegraphics[width=\linewidth]{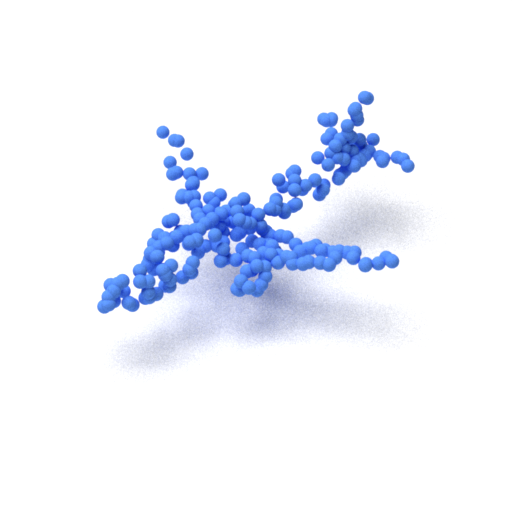}
			\includegraphics[width=\linewidth]{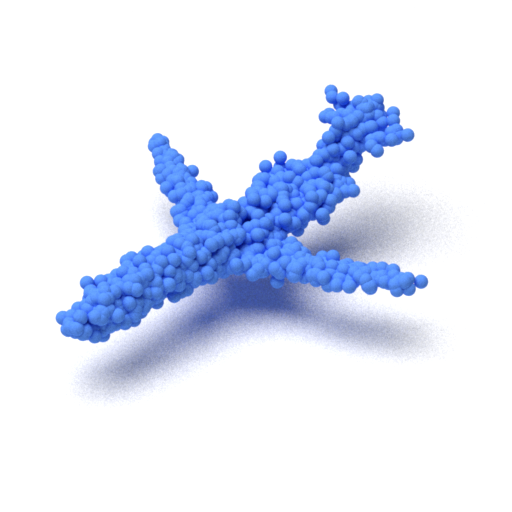}
			\includegraphics[width=\linewidth]{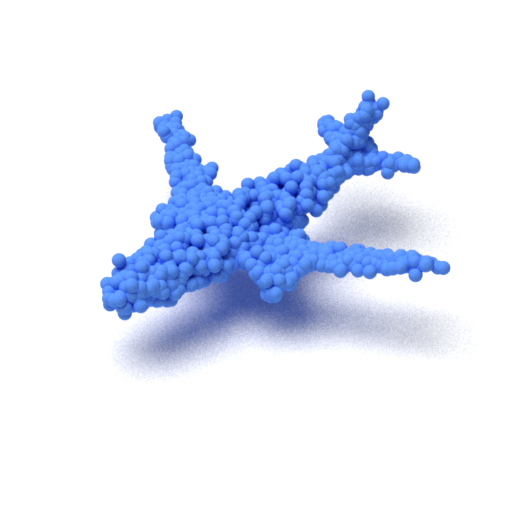}
		\end{subfigure}
		\begin{subfigure}{.3\textwidth}
\includegraphics[width=\linewidth]{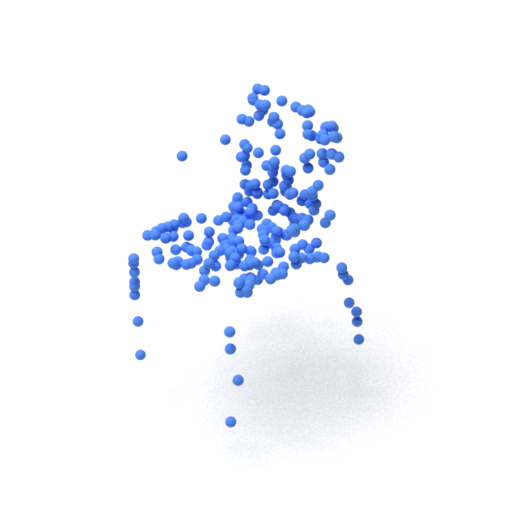}
			\includegraphics[width=\linewidth]{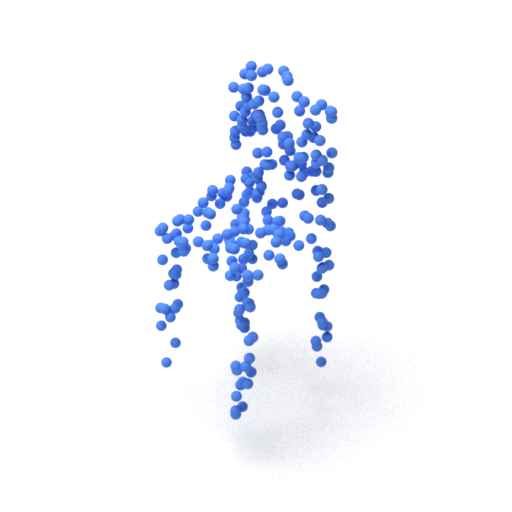}
			\includegraphics[width=\linewidth]{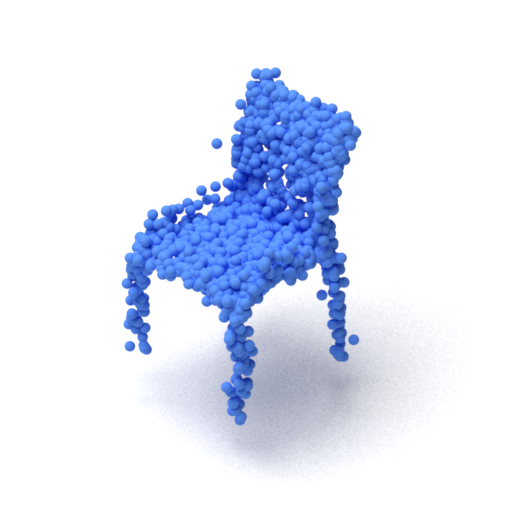}
			\includegraphics[width=\linewidth]{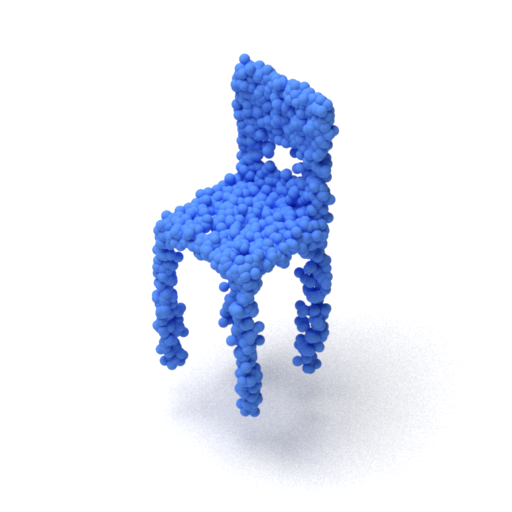}
		\end{subfigure}
		\begin{subfigure}{.3\textwidth}
\includegraphics[width=\linewidth]{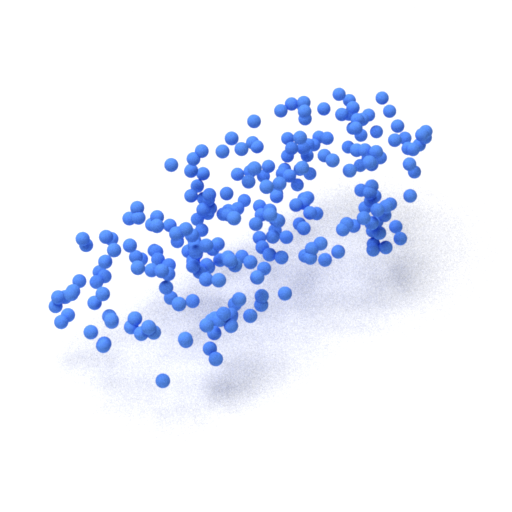}
			\includegraphics[width=\linewidth]{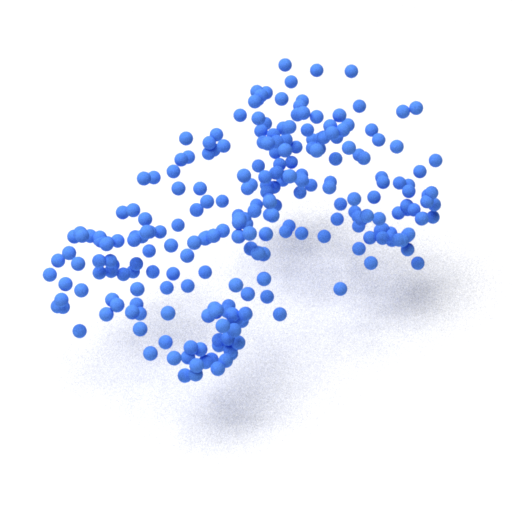}
			\includegraphics[width=\linewidth]{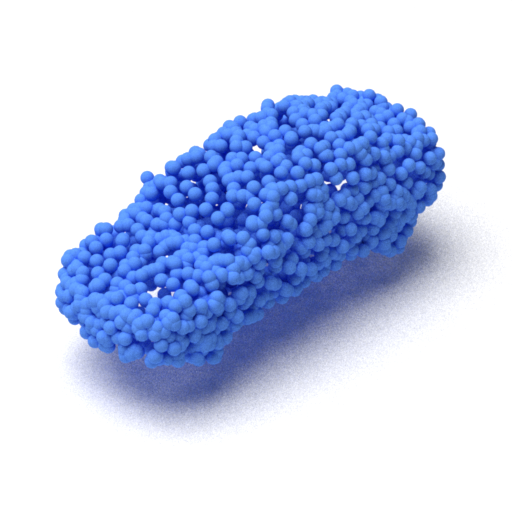}
			\includegraphics[width=\linewidth]{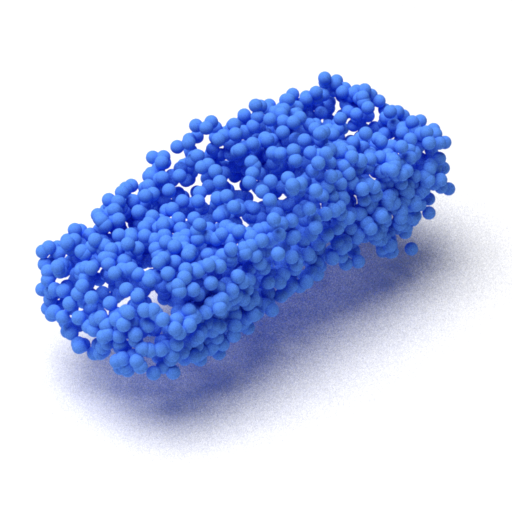}
		\end{subfigure}
		\subcaption{Downsampled point clouds with uniform subsampling~(top) and upsampled
			with the \encoder~(bottom).
		}\label{app:downsample_uniform_appendix}
	\end{subfigure}
	\caption{A visual comparison between down- and upsampling using our
		\downsampler~\subref{app:downsample_downsampler_appendix} and uniform subsampling~\subref{app:downsample_uniform_appendix}.
The downsampled point clouds using the \downsampler are equidistantly
		spread, in contrast to uniform subsampling.
Compare to uniform subsampling, we achieve higher upsampling quality.
Our model has a bias towards equidistant point clouds and leads to a
		better representation of the underlying shape.
}
	\label{app:downsample}
\end{figure}

 \newpage

\section{Additional Experiments}
\label{app:Additional Experiments}

This section provides supplementary figures and tables for the
experiments in the main text. Please note that some results should be
considered \emph{preliminary work}, pointing towards potential future
applications, which we included in the interest of full transparency
regarding our work.

\begin{table*}[h]
	\centering
	\sisetup{
		detect-all              = true,
		table-format            = 2.2(1.2),
		detect-mode             = true,
		separate-uncertainty    = true,
		retain-zero-uncertainty = true,
		mode                    = text,
	}\let\b\bfseries
	\let\i\itshape
	\caption{Reconstruction results for rendering of the three \shapenet classes.
A point cloud is randomly initialised and optimised with the \ipt as a
		loss function.
We repeat this experiment for different resolutions and different
		scales.
The ideal scale is $1/4$ of the resolution~(bottom row).
	}\label{tab:shapenet_render_full}
\resizebox{\linewidth}{!}{\begin{tabular}{cS[table-format=3,drop-zero-decimal=true]
S[table-format=1.2(1.2)]S[table-format=1.2(1.2)]
S[table-format=1.2(1.2)]S[table-format=1.2(1.2)]
S[table-format=2.2(1.2)]S[table-format=1.2(1.2)]
}
			\toprule
			                        & 
			                        & \multicolumn{2}{c}{\emph{Airplane}}
			                        & \multicolumn{2}{c}{\emph{Car}}
			                        & \multicolumn{2}{c}{\emph{Chair}}                                                                                                   \\
			\midrule
			\textsc{Resolution}     & \textsc{Scale Factor}
			                        & {\small CD ($\downarrow$)}
			                        & {\small EMD ($\downarrow$)}
			                        & {\small CD ($\downarrow$)}
			                        & {\small EMD ($\downarrow$)}
			                        & {\small CD ($\downarrow$)}
			                        & {\small EMD ($\downarrow$)}                                                                                                        \\
			\midrule
			\multirow[r]{3}{*}{128} & 128                                 & 1.27 \pm 0.0  & 0.29 \pm 0.0 & 2.97 \pm 0.0  & 0.55 \pm 0.0 & 3.4 \pm 0.01   & 0.65 \pm 0.0  \\
			                        & 64                                  & 0.75 \pm 0.0  & 0.18 \pm 0.0 & 2.15 \pm 0.0  & 0.42 \pm 0.0 & 2.10 \pm 0.0   & 0.41 \pm 0.0  \\
			                        & 32                                  & 0.52 \pm 0.0  & 0.12 \pm 0.0 & 2.04 \pm 0.0  & 0.39 \pm 0.0 & 1.83 \pm 0.0   & 0.35 \pm 0.0  \\\midrule
			\multirow[r]{3}{*}{64}  & 64                                  & 2.65 \pm 0.01 & 0.57 \pm 0.0 & 4.76 \pm 0.01 & 0.85 \pm 0.0 & 5.83 \pm 0.01  & 1.10 \pm 0.0  \\
			                        & 32                                  & 1.88 \pm 0.01 & 0.40 \pm 0.0 & 4.05 \pm 0.01 & 0.70 \pm 0.0 & 4.29 \pm 0.0   & 0.78 \pm 0.0  \\
			                        & 16                                  & 1.22 \pm 0.0  & 0.26 \pm 0.0 & 3.80 \pm 0.0  & 0.64 \pm 0.0 & 3.62 \pm 0.0   & 0.62 \pm 0.0  \\\midrule
			\multirow[r]{3}{*}{32}  & 32                                  & 6.21 \pm 0.01 & 1.32 \pm 0.0 & 9.85 \pm 0.01 & 1.99 \pm 0.0 & 16.92 \pm 0.02 & 3.54 \pm 0.01 \\
			                        & 16                                  & 4.11 \pm 0.01 & 0.90 \pm 0.0 & 8.56 \pm 0.02 & 1.63 \pm 0.0 & 12.55 \pm 0.02 & 2.53 \pm 0.0  \\
			                        & 8                                   & 2.77 \pm 0.0  & 0.61 \pm 0.0 & 7.24 \pm 0.01 & 1.35 \pm 0.0 & 9.05 \pm 0.0   & 1.73 \pm 0.0  \\\bottomrule
		\end{tabular}
	}\end{table*}

 \begin{figure}[!ht]
  \centering
	\begin{subfigure}{.47\textwidth}
		\begin{subfigure}{\linewidth}
			\includegraphics[width=.3\linewidth]{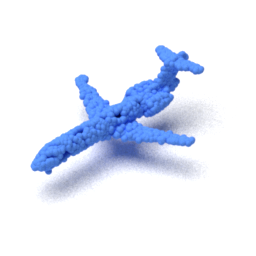}\includegraphics[width=.3\linewidth]{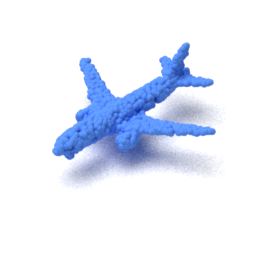}\includegraphics[width=.3\linewidth]{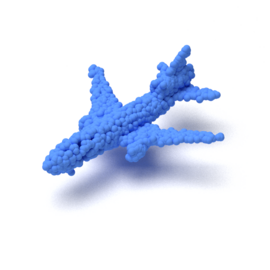}\end{subfigure}
		\begin{subfigure}{\linewidth}
			\includegraphics[width=.3\linewidth]{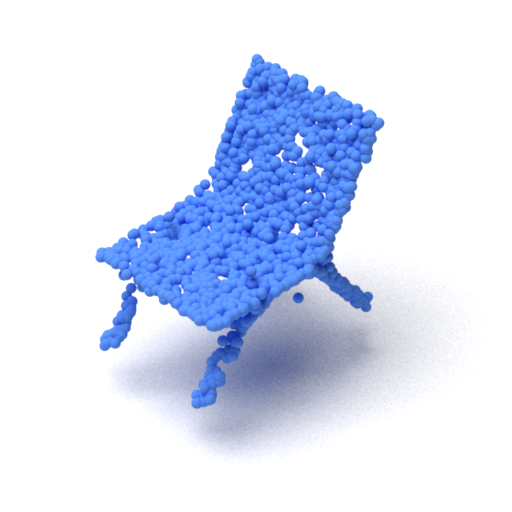}\includegraphics[width=.3\linewidth]{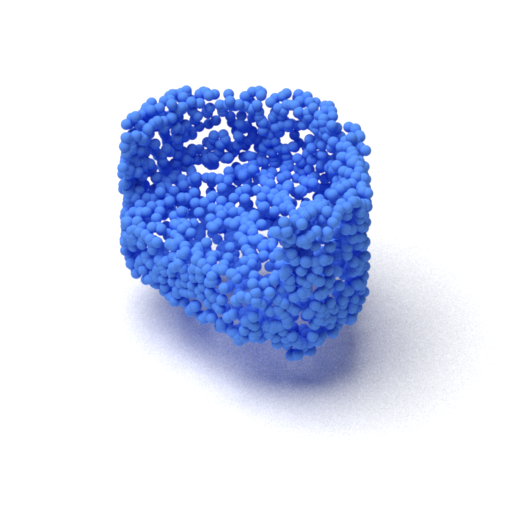}\includegraphics[width=.3\linewidth]{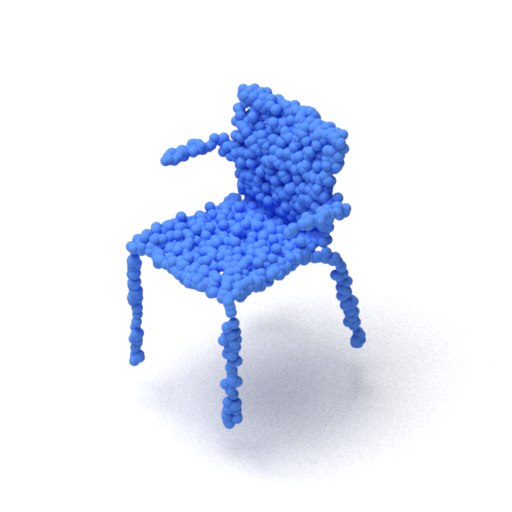}\end{subfigure}
		\begin{subfigure}{\linewidth}
			\includegraphics[width=.3\linewidth]{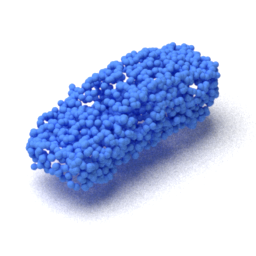}\includegraphics[width=.3\linewidth]{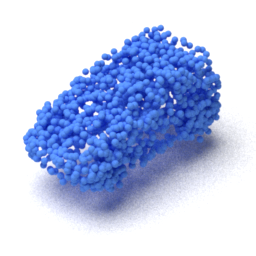}\includegraphics[width=.3\linewidth]{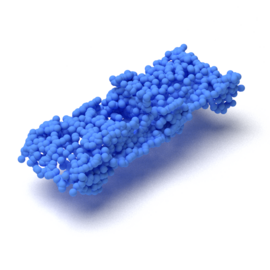}\end{subfigure}
	\end{subfigure}
	\begin{subfigure}{.47\textwidth}
		\begin{subfigure}{\linewidth}
			\includegraphics[width=.3\linewidth]{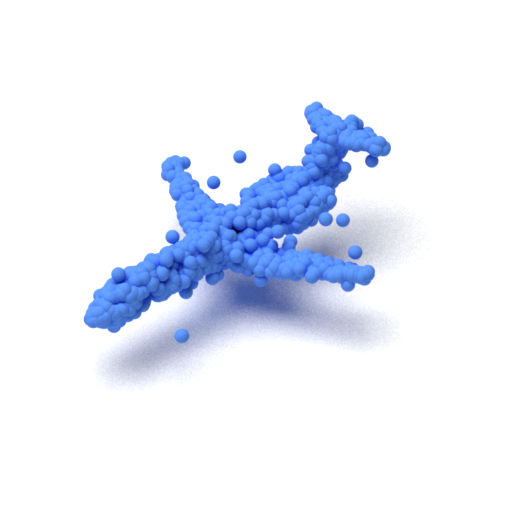}\includegraphics[width=.3\linewidth]{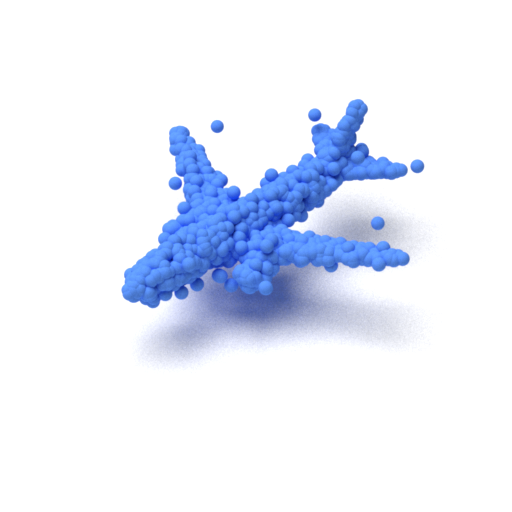}\includegraphics[width=.3\linewidth]{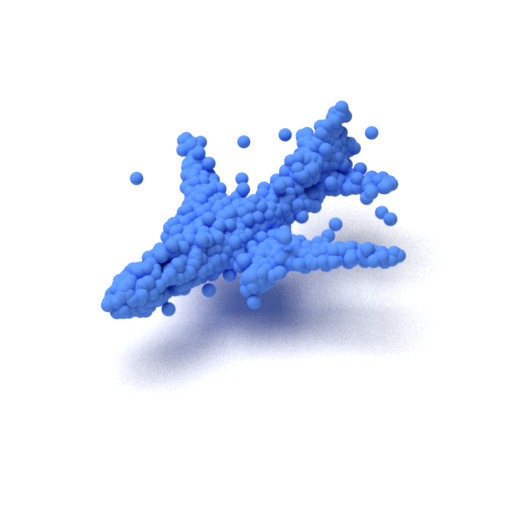}\end{subfigure}
		\begin{subfigure}{\linewidth}
			\includegraphics[width=.3\linewidth]{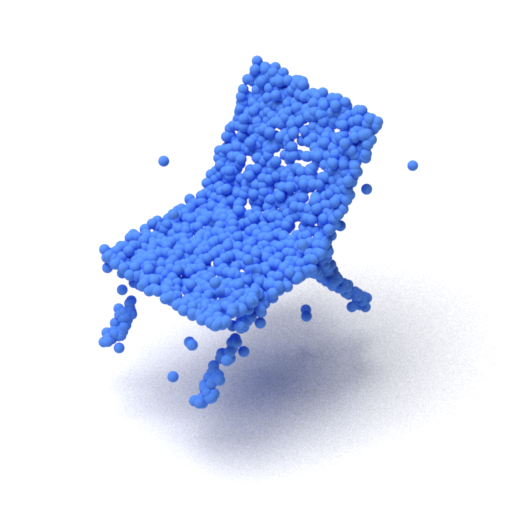}\includegraphics[width=.3\linewidth]{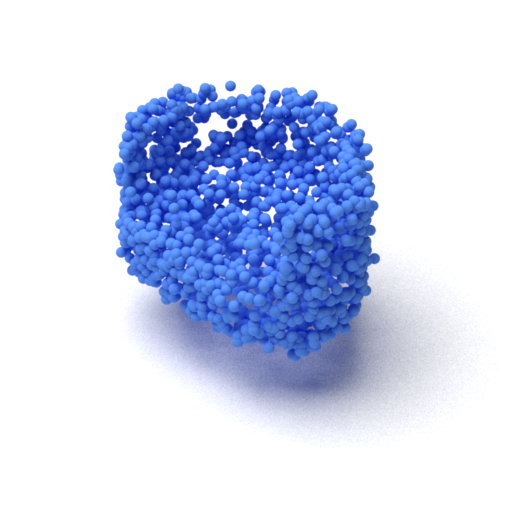}\includegraphics[width=.3\linewidth]{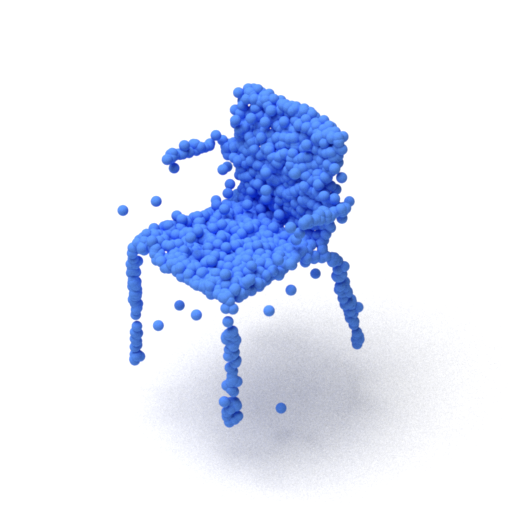}\end{subfigure}
		\begin{subfigure}{\linewidth}
			\includegraphics[width=.3\linewidth]{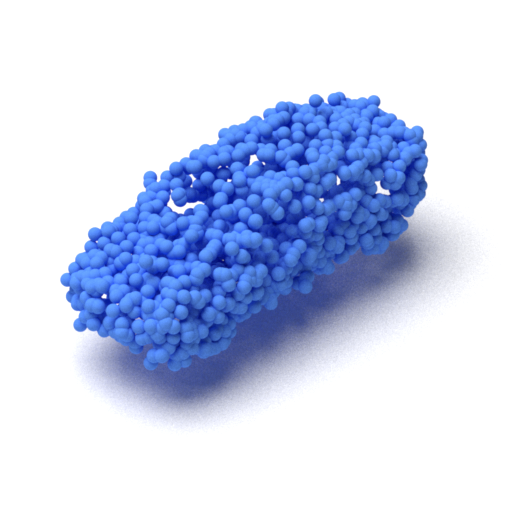}\includegraphics[width=.3\linewidth]{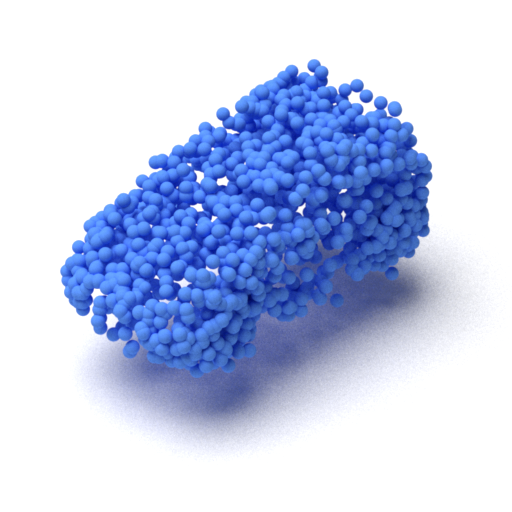}\includegraphics[width=.3\linewidth]{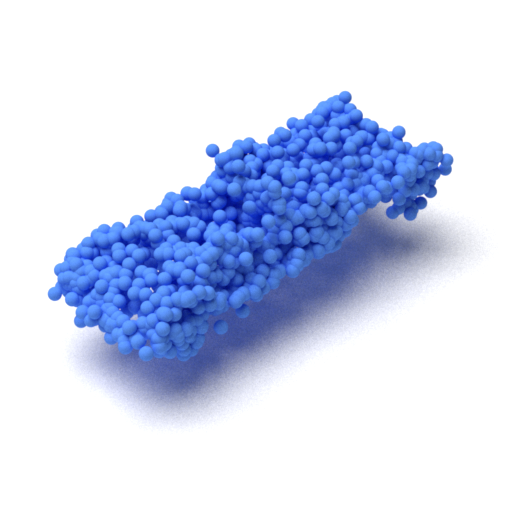}\end{subfigure}
	\end{subfigure}
	\caption{Examples of \emph{rendered} point clouds using our optimisation scheme for the
		three classes of \shapenet.
On the left, the point clouds are rendered at a resolution of $128$ with a scale of $32$, leading
		to high-quality reconstructions.
The right shows the \emph{same} rendering with the scale set to $128$,
		where we observe unstable gradients.
The resulting point clouds have only partially converged, with some
    clear outliers.
	}\label{fig:shapenetcore_rendered}
\end{figure}
 
\begin{table*}[ht]
	\centering
	\sisetup{
		detect-all              = true,
		table-format            = 2.2,
		detect-mode             = true,
		separate-uncertainty    = true,
		retain-zero-uncertainty = true,
		mode                    = text,
	}\caption{Reconstruction of point clouds based on \emph{partial views} using the GenRe~\citep{Zhang18a} dataset.
The point clouds in the dataset are sampled from depth image in a particular direction
		and the task consists of reconstructing the full point from the partial view.
We asses the capability of the \encoder to reconstruct the full point
		cloud from the partial view \emph{without} any additional training.
Thus, the model is only trained on the set of \emph{complete} point clouds.
We use the trained \encoder from \cref{sec:Reconstructing Point
    Clouds}; CD and EMD have to be multiplied by $\num{1e3}$ and
    $\num{1e2}$, respectively.
	}\label{tab:shapenet_completion}
\let\b\bfseries
	\let\i\itshape
	\begin{tabular}{l
S[table-format=2.2]@{\hspace{6pt}}S[table-format=2.2]S[table-format=1.2]@{\hspace{6pt}}S[table-format=1.2]}
		\toprule
		          & \multicolumn{2}{c}{\emph{Chair}{\small\,($\downarrow$)}}
		          & \multicolumn{2}{c}{\emph{Car}{\small\,($\downarrow$)}}                                 \\\midrule
		{\textsc{Model}} & {\small CD }
		          & {\small EMD}
		          & {\small CD }
		          & {\small EMD}                                                                           \\
		\midrule
		SoftFlow  & 2.786                                                    & 3.295   & 1.850   & 2.789   \\
		PointFlow & 2.707                                                    & 3.649   & 1.803   & 2.851   \\
		PVD       & 3.211                                                    & 2.939   & 1.774   & 2.146   \\
		\midrule
		\encoder  & \b 1.10321 & \b 1.23785 & \b .706586 & \b .911902 \\
		\bottomrule
	\end{tabular}
\end{table*}

\begin{table*}[ht]
	\centering
	\sisetup{
		detect-all              = true,
		table-format            = 3.2,
		detect-mode             = true,
		separate-uncertainty    = true,
		retain-zero-uncertainty = true,
		mode                    = text,
	}\caption{Out-of-distribution results for the \encoder model.
The \encoder trained on the class of airplanes is used to reconstruct the category of cars and
		chairs to provide an insight into true out-of-distribution reconstruction and capacity to
		generalise.
Results indicate that our model is not \emph{fully general} and
    thus incapable of reconstructing arbitrary point cloud
    configurations.
Due to the simplicity of our model architecture, this is partially to
    be expected. We hope to improve the multi-modal
    capabilities of our model in future work.
	}\label{tab:Shapenet Cross Reconstruction}
\let\b\bfseries
	\let\i\itshape
\begin{tabular}{l
S[table-format=3.2]@{\hspace{6pt}}S[table-format=2.2]}
		\toprule
		{\textsc{Airplane to}} & {\small CD }
		                       & {\small EMD}          \\
		\midrule
		Car                    & 22.676       & 9.363  \\
		Chair                  & 203.688      & 52.287 \\
		\bottomrule
	\end{tabular}\end{table*}

\begin{table*}[!h]
	\centering
	\sisetup{
		detect-all              = true,
		table-format            = 1.2,
		detect-mode             = true,
		separate-uncertainty    = true,
		retain-zero-uncertainty = true,
		table-align-text-after  = false,
		mode                    = text,
	}\caption{Ablation with respect to the number of directions used in the \ipt to investigate the impact on
		reconstruction quality.
The spatial resolution is kept at $128$~(the default) and we
    progressively reduce the number of directions.
Although the reconstruction quality remains \emph{consistently}
    high, the number of directions has, in general, a positive impact on
    reconstruction quality.
	}\label{tab:shapenet_num_dirs}
\let\b\bfseries
	\let\i\itshape
	\footnotesize \begin{tabular}{S[round-mode=none,table-format=4]@{\hspace{6pt}}S[table-format=1.2]@{\hspace{6pt}}S[table-format=1.2]}
		\toprule
		    & \multicolumn{2}{c}{\emph{Airplane}\,($\downarrow$)}                   \\
		\midrule
		\textsc{Number of directions}
		    & {\small\sc CD}                                      & {\small\sc EMD} \\\midrule
		4   & 1.287                                               & 2.047           \\
		8   & 1.177                                               & 1.882           \\
		16  & 1.090                                               & 1.789           \\
		32  & 1.070                                               & 1.685           \\
		64  & 1.033                                               & 1.559           \\
		128 & 1.05                                                & 1.57            \\
		\bottomrule
	\end{tabular}\end{table*}

\begin{table*}[!h]
	\centering
	\sisetup{
		detect-all              = true,
detect-mode             = true,
}\caption{Additional downsampling experiments.
We follow the same setup as in \cref{sec:Downsampling} and train
		a separate downsampling model for each cardinality.
With each model, we predict a downsampled version of the point
    cloud, which is subsequently upsampled with the \encoder, to obtain a
		reconstruction of the original point cloud.
We observe high quality over all cardinalities, even when the original
		point cloud is summarized with only $32$ points~(!).
As a comparison, we add the original reconstruction result~(cf.\
    \cref{tab:shapenet_reconstruction}) without downsampling in the
    second to last row.
The bottom part of the table demonstrates the capacity of the
    \encoder to reconstruct larger point clouds~(without downsampling).
Although overall performance is still high, our model has the
    \emph{fundamental limitation} that it predicts the point clouds in
    a \emph{single} $N\times 3$ vector.
To generate large point clouds, we thus believe that different
    architectures, such as set transformers~\citep{Lee19a}, will provide
    better scalability with respect to the cardinality.
	}
	\label{tab:shapenet_points_ablation}
\let\b\bfseries
	\let\i\itshape
	\footnotesize \begin{tabular}{S[round-mode=none,table-format=4]@{\hspace{6pt}}S[table-format=1.2]@{\hspace{6pt}}S[table-format=1.2]@{\hspace{6pt}}S[table-format=2.2]@{\hspace{6pt}}S[table-format=2.2]@{\hspace{6pt}}S[table-format=1.2]@{\hspace{6pt}}S[table-format=1.2]@{\hspace{6pt}}}
		\toprule
		     & \multicolumn{2}{c}{\emph{Airplane}\,($\downarrow$)}
		     & \multicolumn{2}{c}{\emph{Chair}\,($\downarrow$)}
		     & \multicolumn{2}{c}{\emph{Car}\,($\downarrow$)}                                                        \\\midrule
		\textsc{Number of points}
		     & {\small\sc CD}
		     & {\small\sc EMD}
		     & {\small\sc CD}
		     & {\small\sc EMD}
		     & {\small\sc CD}
		     & {\small\sc EMD}                                                                                       \\\midrule
		32   & 1.39824                                             & 3.37575 & 18.2513 & 11.6149 & 9.99312 & 6.04725 \\
		64   & 1.19708                                             & 2.6825  & 14.3356 & 9.78847 & 7.57154 & 4.96853 \\
		128  & 1.14626                                             & 2.31211 & 12.0703 & 8.87836 & 6.64424 & 4.50962 \\
		256  & 1.12873                                             & 2.16891 & 11.5905 & 8.53007 & 6.28949 & 4.23256 \\
		512  & 1.12432                                             & 2.09143 & 11.5482 & 8.33557 & 6.31585 & 4.36063 \\
		\midrule
		2048 & 1.05                                                & 1.57    & 9.24    & 6.19    & 5.82    & 3.18    \\
		4096 & 0.867208                                            & 1.60767 & 8.16931 & 6.32711 & 5.13266 & 3.27385 \\
		\bottomrule
	\end{tabular}\end{table*}
 \begin{table*}[t]
	\centering
	\sisetup{
		detect-all              = true,
		table-format            = 2.2,
		detect-mode             = true,
		separate-uncertainty    = true,
		retain-zero-uncertainty = true,
		mode                    = text,
	}\caption{Preliminary results for \emph{multi-class reconstruction} on the
    ShapeNet13 dataset.
The dataset contains 13 classes~(airplanes, cars, chairs, lamps,
    tables, sofas, cabinets, benches, telephones, speakers, monitors,
    vessels, and rifles) from the full ShapeNet dataset.
We follow the experimental setup of \citet{zhou2021pvd} and cite
    their results as comparison partners.
Our model performs well despite its conceptual simplicity; we
    envision more expressive architectures to perform even better.
	}\label{tab:shapenet_multiclass}
\let\b\bfseries
	\let\i\itshape
	\begin{tabular}{lS[table-format=2.2]@{\hspace{6pt}}S[table-format=2.2]}
		\toprule
		                        {\sc\small Model} & {\small CD~($\downarrow$)} & {\small EMD~($\downarrow$)} \\
		\midrule
		                             PVD               & 58.65        & 57.85        \\
		                             PointFlow         & 63.25        & 66.05        \\
		                             LION              & \b 51.85     & \b 48.95        \\
    \midrule
		                             \encoder\,(Ours)  & 63.92        & 50.28        \\
		\bottomrule
	\end{tabular}\end{table*}
 \clearpage
\section{Metrics for Point Clouds}
\label{app:Metrics for Point Clouds}

A good metric for point cloud generation balances computational speed
and theoretical guarantees.
Finding such metrics is a challenging task, since often computations
require the consideration of all pairs of points between the two point
clouds.
An example is the Gromov--Hausdorff distance~\citep{memoli2004comparing},
which has advantageous theoretical properties, but is hard to evaluate.
Two metrics are commonly used to describe the distance between point
clouds, the Chamfer Distance~(CD) and Earth Mover's Distance~(EMD), for
which we present a self-contained summary here.
Although not a metric in the mathematical sense, CD poses a good balance
between computational speed and quality and is defined for point clouds
$X$ and $Y$ as
\begin{equation}
	\text{CD}(X,Y) = \sum_{x\in X} \min_{y\in Y} \|x - y\| + \sum_{y\in Y} \min_{x\in X} \|x-y\|.
\end{equation}
For the EMD, by contrast, the distance between point clouds is viewed as
the cost required to transport one point cloud into the other, i.e.\
\begin{equation}
	\text{EMD}(X,Y) = \min_{\phi:X\to Y} \sum_{x\in X} \| x - \phi(x)\|,
\end{equation}
where $\phi$ solves an optimal transport problem.
Solving the optimal transport problem is a computationally intensive
task that becomes prohibitive for medium to large point clouds.
The properties of using the CD as loss term were investigated in
\citet{achlioptas2018learning}, revealing that reconstructions had
non-uniform surface density, compared to the uniformly sampled ground
truth points.
A strong advantage of the CD is its computational efficiency for medium
to large point cloud cardinalities.

\newpage

\section{Architectural Details}
\label{app:Architectural Details}
We provide the details of both the \encoder and \vae architecture.
The \ipt as defined in \cref{eq:IPTIndicator} is viewed as an image of
shape $n_{h} \times n_{d}$, where $n_{h}$ is the discretisation of the
heights along each direction and $n_{d}$ is the number of directions.
In the architecture, $n_{d}$ denotes channels in the CNN; to
accommodate for this, we expect the input to be of the form $n_{b}\times
	n_{d}\times n_{h}$, where $n_{b}$ is the batch size.
In all our experiments we set $n_{d}=n_{h}=128$ and denote a batch of
\ipt{}s in the diagram with \texttt{IPT[B,128,128]}.
The internal architecture of the \encoder is:
\begin{center}
	\ttfamily
	\begin{tabular}{c}
		IPT[B,128,128]                        \\
		|                                     \\
		3x[Conv1d-BatchNorm1d-SiLU-MaxPool1d] \\
		|                                     \\
		Conv1d                                \\
		|                                     \\
		Flatten                               \\
		|                                     \\
		FC-ReLU-FC                            \\
		|                                     \\
		Tanh-FC                               \\
		|                                     \\
		PointCloud[B,2048,3]
	\end{tabular}
\end{center}
In the architecture above, \texttt{PointCloud[B,2048,3]} denotes the
\emph{final} predicted batch of point clouds.

Our \vae has the following architecture:
\begin{center}
	\ttfamily
	\begin{tabular}{c}
		IPT[B,128,128]                          \\
		|                                       \\
		4x[Conv1d-LayerNorm-ReLU]               \\
		|                                       \\ 
		Conv1d                                  \\
		|                                       \\
		\{FC-MU,FC-VAR\}                        \\
		|                                       \\
		Latent Space                            \\
		|                                       \\
		FC                                      \\
		|                                       \\
		4x[Conv1DTranspose-BatchNorm-LeakyReLU] \\
		|                                       \\
		\verb|[|Conv1D-Tanh\verb|]|             \\
		|                                       \\
		IPT[B,128,128]
	\end{tabular}
\end{center}
In the diagram above, latent embeddings are denoted by \texttt{FC-MU}
and \texttt{FC-VAR}, respectively.
\newpage

\section{Properties of the Inner Product Transform}
\label{sec:Inner Product Transforms}

This section discusses all proofs for statements from the main paper.
While the first result about the injectivity is already known, we
provide a novel, highly-accessible proof, which only requires basic
concepts from linear algebra.
In the following, we will consider two point clouds $X,Y\subset
	\mathbb{R}^{n}$ to be equal if they are equal in the sense of sets; in
particular, this implies that they need to have the same cardinality.
We will also assume that all points are in \emph{general position}.
\iptinjective*
\begin{proof}
	Let $X,Y\subset \mathbb{R}^{n}$ with $X \neq Y$ and let $\Xi \subset
		S^{n-1}$ be a set of $n+1$ affinely-independent directions.
Given $\xi\in\Xi$, we write $X_{\xi}:=\{ \langle x, \xi \rangle \mid
		x\in X\}$ for the projection of $X$ onto the one-dimensional
	subspace along $\xi$, with $\langle x, \xi \rangle $ denoting the
	standard Euclidean inner product. For $t \in \mathbb{R}$, we define
	$X_{\xi, t} := \{x \in X \mid \langle x, \xi \rangle \leq t\}$.
If for any direction $\xi\in \Xi$ we have $X_{\xi} \neq Y_{\xi}$, we
	are done because this means that we can find thresholds $t_1, t_2 \in
		\mathbb{R}$ with $t_1 < t_2$ such that for $t \in [t_1, t_2]$, the
	cardinality of $X_{\xi, t}$ changes but the cardinality of $Y_{\xi,
				t}$ does not change or vice versa.
Thus, let us assume that $X_{\xi} = Y_{\xi}$ for all directions
	$\xi$.
For each $\xi$, we may sort the values by magnitude and calculate
	differences, i.e.\ expressions of the form $\langle x, \xi\rangle
		- \langle y, \xi\rangle$.
We have $\langle x, \xi\rangle - \langle y, \xi\rangle = 0$ for all
	directions $\xi \in \Xi$ by assumption, which we may rewrite as
	$\langle x - y, \xi \rangle = 0$.
Treating this as a system of $n+1$ linear equations, this is
	equivalent to stating that the kernel of the corresponding linear
	map is the whole domain.
However, since the $n+1$ directions are affinely independent, all
	coefficients must be zero, implying that the point clouds are the
	same.
This is a contradiction, so our initial assumption must have been
	wrong. Thus, there is $\xi \in \Xi$ such that $X_\xi \neq Y_\xi$,
	so we have $\mathrm{\ipt}(X) \neq \mathrm{\ipt}(Y)$.
\end{proof}

\iptmetric*
\begin{proof}
If $d(X,Y)=0$ then $\Vert \text{IPT}(X) - \text{IPT}(Y) \Vert_{2} = 0$ and
	since $\Vert\cdot\Vert_{2}$ is a metric, it follows that $\text{IPT}(X) = \text{IPT}(Y)$.
	Since the \ipt is injective, we conclude that $X=Y$.
Equality in this case is seen as equality the of sets, that is up to
	permutation.
Both the reverse implication and symmetry follow from the definitions.
For the triangle inequality, we verify
	\begin{equation}
		\begin{aligned}
			d(X,Y) & = \Vert \text{IPT}(X) - \text{IPT}(Y) \Vert_{2}                                             \\
			       & = \Vert (\text{IPT}(X) - \text{IPT}(Z)) - (\text{IPT}(Y)- \text{IPT}(Z)) \Vert_{2}          \\
			       & \leq \Vert \text{IPT}(X) - \text{IPT}(Z)\Vert + \Vert\text{IPT}(Y)- \text{IPT}(Z) \Vert_{2} \\
			       & = d(X,Z) + d(Z,Y).
		\end{aligned}
	\end{equation}
\end{proof}
\iptlinear*
\begin{proof}
	It follows from the definition of the \ipt that for all $\xi\in S^{n-1}$ and
	$h\in\mathbb{R}$ we have
	\begin{equation}
		\begin{aligned}
			\textrm{IPT}(X\cup Y) & = \sum_{x\in X\cup Y} \boldone_{x}(\xi,h)                               \\
			                      & = \sum_{x\in X} \boldone_{x}(\xi,h) + \sum_{y\in Y} \boldone_{y}(\xi,h) \\
			                      & = \textrm{IPT}(X) + \textrm{IPT}( Y).
		\end{aligned}
	\end{equation}
The second equality uses that the intersection of $X$ and $Y$ is
	empty, implying that a point is either in $X$ or in $Y$.
\end{proof}
\iptsurjective*
\begin{proof}
	To show surjectivity, it suffices to show that for any rational
	linear combination of \ipt{}s there exists a point cloud
	$Z\subset\mathbb{R}^n$ that has an IPT equal to that linear
	combination, up to a rational coefficient.
This is to say that
	$\forall \alpha,\beta \in \mathbb{Q} \exists \gamma \in \mathbb{Q}$ such that
	\begin{equation}
		\alpha\textrm{IPT}(X) + \beta \textrm{IPT}(Y) = \gamma \textrm{IPT}(Z).
	\end{equation}

	Let $\alpha = p / q$ and $\beta = r / s$, then
	\begin{equation}
		\begin{aligned}
			qs \bigg[ \frac{p}{q}\textrm{IPT}(X) + \frac{r}{s}\textrm{IPT}(Y)      \bigg] & = sp\textrm{IPT}(X) + qr\textrm{IPT}(Y) \\
			                                                                              & = \textrm{IPT}(\cup_{sp}X \cup_{qr}Y).
		\end{aligned}
	\end{equation}
	Thus, setting $\gamma = 1/qs$ and $Z = \cup_{sp}X \cup_{qr}Y $ does
	the trick.
We conclude that the \ipt is surjective on rational linear combinations.
Let $0\leq p\leq q$, statement follows from the equalities
	\begin{equation}
		\begin{aligned}
			\frac{p}{q}\textrm{IPT}(X) & + \frac{q-p}{q}\textrm{IPT}(Y)                                                                 \\
			                           & = \frac{1}{q}\big[ p\, \textrm{IPT}(X) + (q-p)\, \textrm{IPT}(Y)\big]                          \\
			                           & = \frac{1}{q}\left[\textrm{IPT}\big(\cup_{p} X\big) + \textrm{IPT}\big(\cup_{q-p}Y\big)\right] \\
			                           & = \frac{1}{q}\textrm{IPT}\big( \cup_{p} X  \cup_{q-p} Y \big).
		\end{aligned}
	\end{equation}
\end{proof}
 \clearpage

\end{document}